\documentclass[]{elsarticle}

\makeatletter
\def\ps@pprintTitle{%
 \let\@oddhead\@empty
 \let\@evenhead\@empty
 \def\@oddfoot{LLNL-JRNL-833876-DRAFT}%
 \let\@evenfoot\@oddfoot}
\makeatother

\usepackage{bm}
\usepackage{caption}
\usepackage{subcaption}
\usepackage{amsmath}
\usepackage{amssymb}
\usepackage{lineno,hyperref}
\usepackage{xcolor}

\modulolinenumbers[5]
\newcommand{\age}{\textsf{Age}}
\newcommand{\sex}{\textsf{Sex}}
\newcommand{\death}{\textsf{Death}}
\newcommand{\beds}{\textsf{Beds}}
\newcommand{\adm}{\textsf{Admission Diagnoses}}
\newcommand{\dis}{\textsf{Discharge Diagnoses}}
\newcommand{\labs}{\textsf{Laboratory Tests}}
\newcommand{\neuro}{\textsf{Neurological Tests}}
\newcommand{\meds}{\textsf{Medications}}

\begin{document}

\begin{frontmatter}
\title{Unsupervised Probabilistic Models for Sequential Electronic Health Records}

\author[llnl]{Alan D. Kaplan}
\author[kp]{John D. Greene}
\author[kp]{Vincent X. Liu}
\author[llnl]{Priyadip Ray}

\address[llnl]{Computational Engineering Division, Lawrence Livermore National Laboratory, 7000 East Ave., Livermore, CA 94550}
\address[kp]{Kaiser Permanente Division of Research, 2000 Broadway, Oakland, CA 94612}

%\affiliation[llnl]{organization={Lawrence Livermore National Laboratory},
%        addressline={7000 East Avenue},
%        city={Livermore},
%        postcode={94550},
%        state={CA},
%        country={USA}}
%        
%\affiliation[kp]{organization={Kaiser Permanente},
%        addressline={},
%        city={},
%        postcode={},
%        state={CA},
%        country={USA}}

%\author{\name Alan D. Kaplan \email kaplan7@llnl.gov \\
%       \addr Computational Engineering Division\\
%       Lawrence Livermore National Laboratory\\
%       Livermore, CA 94550, USA
%       \AND
%       \name Priyadip Ray \email XXXX@XXXX.XXX \\
%       \addr Computational Engineering Division\\
%       Lawrence Livermore National Laboratory\\
%       Livermore, CA 94550, USA
%       \AND
%       \name Vincent X. Liu \email XXXX@XXXX.XXX \\
%       \addr XXXX XXXX \\
%       XXXX XXXX \\
%       XXXX, XX, XXXXX
%       \AND
%       \name John D. Greene \email XXXX@XXXX.XXX \\
%       \addr XXXX XXXX \\
%       XXXX XXXX \\
%       XXXX, XX, XXXXX}
%
%\editor{XXXX XXXX}
%
%\maketitle

\begin{abstract}%   <- trailing '%' for backward compatibility of .sty file
We develop an unsupervised probabilistic model for heterogeneous Electronic Health Record (EHR) data.
Utilizing a mixture model formulation, our approach directly models sequences of arbitrary length, such as medications and laboratory results.
%The model captures associations between sequences, such as medications and laboratory results, that have arbitrary lengths.
This allows for subgrouping and incorporation of the dynamics underlying heterogeneous data types.
The model consists of a layered set of latent variables that encode underlying structure in the data.
These variables represent subject subgroups at the top layer, and unobserved states for sequences in the second layer.
We train this model on episodic data from subjects receiving medical care in the Kaiser Permanente Northern California integrated healthcare delivery system.
The resulting properties of the trained model generate novel insight from these complex and multifaceted data.
In addition, we show how the model can be used to analyze sequences that contribute to assessment of mortality likelihood.
\end{abstract}

%\begin{keywords}
%  Electronic Health Records, Latent Variable Models, Machine Learning, Unsupervised Learning, Sequence Models
%\end{keywords}

\end{frontmatter}
%\linenumbers

\section{Introduction}
The use of Electronic Health Record (EHR) data collected during routine clinical care is critical to the aspirations of precision medicine \cite{Ginsburg2018-zf, Kosorok2019-bz}.
EHR repositories contain large amounts of wide-ranging patient and treatment information and are essential for the development of individualized treatments in the context of disease progression \cite{Kim2019-bo}.
%Beyond genomics, the idea is that hidden within collections of Electronic Health Record (EHR) data, lie patterns that may assist in the formulation of individualized medical decision making.
With the broad adoption of EHR in the US, a large variety of data types are now routinely collected over long periods of time.
%Given the array of problems that can be addressed with EHRs, it makes sense to widen the search in case valuable information may be hidden in hitherto unknown locations.
This has ushered in an era of research focused on the applications and development of data-analytic tools for mining historical records of medical data to drive novel insight.
Broadly, the extraction of meaningful patterns through unsupervised learning \cite{Pivovarov2015-zu, Mayhew2018-pt, Li2016-vp, Huang2018-oy} and the prediction of outcomes through supervised learning \cite{Luz2020-rp, Zhou2016-po, Xie2020-jl, Wu2010-hr, Su2020-sq, Levine2018-ih, Kam2017-tz, Rasmy2018-gi, Barak-Corren2017-ax} are two important directions.

%The goal of unsupervised learning is to model the data without specific prediction targets defined a priori.
Unsupervised methods can be applied towards many different tasks, such as prediction, imputation, and simulation; and 
often contain a model of the underlying structure in the data \cite{Murphy2012-kt}.
%can be used to reveal underlying structure in the data \cite{Murphy2012-kt}.
This underlying structure is not directly observed and can lead to insights that are otherwise difficult to produce, especially for large and complex data sets.
%be difficult to identify when faced with heterogeneous data.
Subgroup identification is one particular aspect of unsupervised learning that is of high relevance to precision medicine, since subgroups are often a driving aspect of treatment planning.
In particular, probabilistic versions of unsupervised learning models
are appealing because they contain a representation of the uncertainty attached to subgroup identification \cite{Ghahramani2015-xn}.
%carry  appeal in that they contain a representation of the uncertainty attached to 
%inference and prediction \cite{Ghahramani2015-xn}.

%In many applications of clinical data analysis, a specific prediction target may not be known, or available.
%In other cases there may be multiple ways of measuring a single outcome, each one having certain advantages and disadvantages.
%More commonly, there may be multiple outcomes of interest.
%These facets of clinical applications lead to increased emphasis on unsupervised learning, where the goal is to model the data without specific prediction targets defined a priori.
%From unsupervised models, we seek to learn about the structure of the data, using the models to infer likely outcomes, and generate hypothesis that could be further tested \cite{Murphy2012-kt}.
%Probabilistic versions of unsupervised learning carry additional appeal in that
%%, once trained, arbitrary conditional probabilities can be computed 
%they contain a representation of the uncertainty attached to inference and prediction \cite{Ghahramani2015-xn}.
%This flexibility is especially useful in the application areas outlined above.

The goal of this work is to develop a general model capable of uncovering patterns within EHR sequences, with respect to learned subgroups.
The probabilistic formulation leads to the ability to compute likelihoods of subgroup membership and likely EHR sequences.
In this work, we explore the trained model with respect to its learned subgroups.
This model can be used in many different applications, including:
\begin{itemize}
    \item \emph{Disease phenotyping}. Given a definition for disease (such as a diagnosis code, or combination of codes), a trained model could be used to generate sequence-level phenotypes.
    \item \emph{Resource allocation}. Since the model includes the likelihood of the length of sequences, it is possible to infer the length of, for example, the Beds sequence given partial values of other sequences. This could be used to estimate future demand for hospital resources (e.g., ICU beds).
    \item \emph{Prediction}. For a specific application, outcome definitions could be defined at discharge-time and included as variables in the model. Then these variables can be inferred probabilistically using the model, given input sequence data
\end{itemize}

However, EHR data do not readily conform to rectangular shapes required by many of these modeling approaches.
%come in a variety of shapes and sizes.
Common characteristics of EHR include non-uniformly sampled and variable length sequences, known as heterogeneous data \cite{Wang2017-iz}.
This includes, for example, collections of vital signs or laboratory results.
Many existing machine learning approaches require data to be rectangular in shape, and therefore transforming the data to fit existing methods is one avenue for predictive modeling \cite{Shickel2018-qy}.
One way of accomplishing this would be to summarize sequence statistics in binned time intervals.
A difficulty with this approach is selecting the width of time window and the loss of granularity within each window.
In summary, information can be lost in transformation.

Mixture models are flexible and powerful latent variable formulations that provide many benefits of unsupervised learning \cite{McLachlan2004-mh}.
Of clinical importance, the latent variable probabilistically associates individuals with one of several subgroups that are learned from the data.
Methods based on the mixture model have been investigated in many applications using EHR data \cite{Najjar2015-rb, Wang2011-zm, Cheung2017-gk, Liu2016-mw, Hubbard2017-hs}.
%are flexible tools that provide many benefits of unsupervised learning.
%Since EHR measurements are often repeated over the course of an illness or treatment, tracking subjects' condition over time is of great importance.
While this provides a solid foundation, it does not inherently provide a way to track changes occurring over time.
%One approach for incorporating temporal dynamics in various domains
%%that utilize mixture models 
%%for tracking the temporal evolution of the subject's state 
%%is to view each sample of the latent variable as a snapshot in time (see e.g. \cite{Briand2015-hw, Mayhew2018-pt}).
%Then these latent temporal snapshots can be tracked as they progress over time.
%However, this approach does not take into account sequential order of the underlying EHR sequences.

Dynamic systems models, such as the Hidden Markov Model (HMM), or predictive methods such as Recurrent Neural Networks (RNN) could address this limitation by directly incorporating temporal dynamics.
HMM methods can be formulated as continuous-time models, where transition probabilities depend on  inter-measurement intervals \cite{Stella2012-vr, Liu2015-ad}.
Gaussian Process (GP) models capture temporal dynamics and have been developed for EHR data \cite{Futoma2017-os, Alaa2017-nd, Meng2021-uy, Li2021-tw}.
%However, GP methods do not handle heterogeneous data types, including categorical items and simultaneous observations (i.e., multiple medications administered).
RNN approaches can be used by predicting 
%A method utilizing RNN structures for prediction 
the time until an event from the current input and the time since the last input \cite{Choi2016-pr}.
Versions of the RNNs, such as 
Long Short-Term Memory (LSTM) for capturing long-term dynamics have also been applied to EHR data \cite{Jin2018-rh}.
%for predicting heat failure from a sequence of diagnosis codes has been investigated \cite{Jin2018-rh}.
%Probabilistic methods based on Markov properties include continuous-time versions that capture the distribution of inter-measurement intervals \cite{Stella2012-vr, Liu2015-ad}.
However, these dynamic predictive approaches do not offer the ease of subgroup interpretation that the mixture model affords.

Methods that combine mixture model structures allowing for subgrouping and dynamic modeling include mixtures of HMM and mixtures of GP models \cite{Rasmussen2001-wh, Meeds2005-zd}.
This includes a model for patient risk scoring \cite{Alaa2018-nq}, clustering gene expression data \cite{McDowell2018-ae} and COVID-19 trajectory subgrouping \cite{Cui2022-pn}.
Several methods have been developed for application areas outside of the biomedical domain \cite{Yuksel2012-iq, Piyathilaka2013-dr}.
%These have been incorporated in approaches for detecting sepsis [futoma] DOING X and Y CITE.
Although these methods allow for subgrouping, they do not handle heterogeneous data types, including categorical items and simultaneous observations (i.e., multiple medications administered).
This heterogeneous quality of EHR of data types is an important aspect that our model is designed to work with.
In our approach, we develop a model that: 1) utilizes mixtures for probabilistic subgroup analysis, 2) captures temporal dynamics of data elements, and 3) conforms to the heterogeneity of data types.

Our method combines both the mixture model and temporal dynamics into one model, 
%Our method takes a different approach, 
where entire sequences are modeled simultaneously as observations in the mixture model.
This leads to each latent state representing an entire temporal trajectory.
%In this work, we develop probabilistic models for episodic EHR data that contain sequences of arbitrary length.
%The model is defined by a full joint probability distribution over the data components and are designed to fit the data without utilizing any lossy transformations.
A layered set of latent variables are used that express membership across population subgroups and categorize observed data elements.
As in the mixture model approach, the top-layer latent variable is an indicator that assigns subjects to one of a set of groupings.
In addition to grouping by subjects, the second layer of latent variables categorizes each data type.
These latent variables are indicators for sequence categories.
This results in a set of states for each data type (\meds{}, \labs{}, etc.), each one  describing probabilities for all possible sequences of the associated data type.
In this way, we decompose the dataset simultaneously across subject states and data states.

\section{Data} \label{sec:data}
Kaiser Permanente Northern California (KPNC) is a highly integrated healthcare delivery system with 21 medical centers caring for an overall population of 4 million members.
For this work, we use a dataset consisting of 244,248 in-patient hospitalization visits with a suspected or confirmed infection and sepsis diagnosis, drawn from KPNC medical centers between 2009 and 2013 \cite{Liu2017-al}.
There can be multiple episodes per subject. There were 203,357 subjects in this dataset. The median number of stays per subject is 1, the mean is 1.6 and the standard deviation is 1.4. The range of the number of episodes per subject is 1 to 65.

We consider time-dependent data elements that appear along a subject's timeline.
Upon hospitalization (at time 0), \age{}, \sex{}, and \adm{} are collected.
The \adm{} are ICD9 codes that are assigned early in the hospitalization, typically at the transition between emergency department and inpatient care.
%assigned initially before any tests are performed or medications are given.
As the patient moves through the hospital, we record the sequence of \textsf{Beds} that they are admitted to.
We capture sequences of both \labs{} and \neuro{}, in which multiple tests may occur at the same timepoint.
From the \labs{}, we extract the test type, but not the test result.
%the ICD9 code for the test, but not the test result.
The \neuro{} consist of neurological evaluations and mental status nursing-based flowsheet assessments including relevant assessment (e.g., Glasgow Coma Score) and the result.

\meds{} are recorded throughout the episode and are categorized by therapy class (a grouping of specific medications).
As in the case of \labs{} and \neuro{}, multiple instances may be given at a single timepoint.
For this work, we did not extract the dosage or frequency of usage.

Table \ref{tbl:seqlen} shows the minimum, mean, standard deviation and maximum number of items in each sequence.
The \labs{}, \meds{}, and \neuro{} can have multiple items per timepoint.
Table \ref{tbl:subseqlen} shows the same statistics for each administration of these categories.

\begin{table}[ht] 
    \centering
     \begin{tabular}{|c c c c c|} 
     \hline
     Sequence & Min & Mean & Std & Max\\ [0.5ex] 
     \hline
     \beds{}        & 1 & 2.83  & 1.68  & 60   \\ 
     \adm{}         & 0 & 13.63 & 8.12  & 90   \\
     \dis{}         & 0 & 8.81  & 6.87  & 143  \\
     \labs{}        & 0 & 12.73 & 17.89 & 964  \\
     \neuro{}       & 0 & 27.19 & 58.35 & 8729 \\  
     \meds{}        & 0 & 15.68 & 10.18 & 193  \\
     \hline
     \end{tabular}
     \caption{Length of Sequences}
     \label{tbl:seqlen}
\end{table}

\begin{table}[ht] 
    \centering
     \begin{tabular}{|c c c c c|} 
     \hline
     Sequence & Min & Mean & Std & Max\\ [0.5ex] 
     \hline
     \labs{}        & 1 & 14.50 & 11.65 & 102 \\
     \neuro{}       & 1 & 2.84  & 1.74  & 13  \\  
     \meds{}        & 1 & 1.31  & 0.84  & 18  \\
     \hline
     \end{tabular}
     \caption{Number of Items per Timepoint}
     \label{tbl:subseqlen}
\end{table}

Notation used for these data streams associated with each episode are shown in Table \ref{tbl:datnot}.
Bold symbols indicate vectors of values.
In cases where we refer to any of these streams, we use the symbol $\boldsymbol{x}$.
The number of values in $\boldsymbol{x}$ is $|\boldsymbol{x}|$.
Note that for \textsf{Laboratory Tests}, \textsf{Neurological Tests}, and \textsf{Medications}, more than one item is possible at each time point.
This means that in general $\boldsymbol{\lambda}, \boldsymbol{\nu}$, and $\boldsymbol{\mu}$  contain sequences of varying length.
%For these sequences, the time associated with a value is denoted $t_{x, i}$.
%For example, the timestamp (in hours) of the collection of laboratory tests $\boldsymbol{\lambda_i}$ is $t_{\lambda, i}$ and the timestamp for the medications $\boldsymbol{\mu_i}$ is $t_{\mu, i}$.
%The \textsf{Age}, \textsf{Sex}, and \textsf{Death} variables do not have timestamps associated with them.
The complete collection of data for one episode is: $\boldsymbol{y}=\left(\phi_a, \phi_s, \phi_d, \boldsymbol{\beta}, \boldsymbol{\alpha}, \boldsymbol{\delta}, \boldsymbol{\lambda}, \boldsymbol{\nu}, \boldsymbol{\mu} \right)$.
Table \ref{tbl:permvals} shows permissible values for each of these sequences.

\begin{table}[]
    \centering
     \begin{tabular}{|c c|} 
     \hline
     Stream & Notation \\
     \hline
     \age{}     & $\phi_a$ \\
     \sex{}     & $\phi_s$ \\
     \death{}   & $\phi_d$ \\
     \beds{}    & $\boldsymbol{\beta} = \{\beta_1, \ldots, \beta_{|\boldsymbol{\beta}|}\}$  \\ 
     \adm{}     & $\boldsymbol{\alpha} = \{\alpha_1, \ldots, \alpha_{|\boldsymbol{\alpha}|} \}$ \\
     \dis{}     & $\boldsymbol{\delta} = \{\delta_1, \ldots, \delta_{|\boldsymbol{\delta}|} \}$ \\
     \labs{}          & $\boldsymbol{\lambda} = \{\boldsymbol{\lambda}_1, \ldots, \boldsymbol{\lambda}_{|\boldsymbol{\lambda}|}\} = 
    \{
    \{\lambda_{1, 1}, \ldots, \lambda_{1, |\boldsymbol{\lambda}_1|} \},
    \ldots,
    \{\lambda_{|\boldsymbol{\lambda}|, 1}, \ldots, \lambda_{|\boldsymbol{\lambda}|, |\boldsymbol{\lambda}_{|\boldsymbol{\lambda}|}|} \}
    \}$ \\
     \neuro{}        & $\boldsymbol{\nu} = \{\boldsymbol{\nu}_1, \ldots, \boldsymbol{\nu}_{|\boldsymbol{\nu}|}\} = 
    \{
    \{\nu_{1, 1}, \ldots, \nu_{1, |\boldsymbol{\nu}_1|} \},
    \ldots,
    \{\nu_{|\boldsymbol{\nu}|, 1}, \ldots, \nu_{|\boldsymbol{\nu}|, |\boldsymbol{\nu}_{|\boldsymbol{\nu}|}|} \}
    \}$ \\  
     \meds{}               & $\boldsymbol{\mu} = \{\boldsymbol{\mu}_1, \ldots, \boldsymbol{\mu}_{|\boldsymbol{\mu}|}\} = 
    \{
    \{\mu_{1, 1}, \ldots, \mu_{1, |\boldsymbol{\mu}_1|} \},
    \ldots,
    \{\mu_{|\boldsymbol{\mu}|, 1}, \ldots, \mu_{|\boldsymbol{\mu}|, |\boldsymbol{\mu}_{|\boldsymbol{\mu}|}|} \}
    \}$ \\
     \hline
     \end{tabular}
     \caption{Notation for data streams.}
     \label{tbl:datnot}
\end{table}

%\begin{itemize}
%    \item \textsf{Age}: $\phi_a$
%    \item \textsf{Sex}: $\phi_s$
%    \item \textsf{Death}: $\phi_d$
%    \item \textsf{Beds}: $\boldsymbol{\beta} = \{\beta_1, \ldots, \beta_{|\boldsymbol{\beta}|}\}$
%    \item \textsf{Admission Diagnoses}: $\boldsymbol{\alpha} = \{\alpha_1, \ldots, \alpha_{|\boldsymbol{\alpha}|} \}$
%    \item \textsf{Discharge Diagnoses}: $\boldsymbol{\delta} = \{\delta_1, \ldots, \delta_{|\boldsymbol{\delta}|} \}$
%    \item \textsf{Laboratory Tests}: $\boldsymbol{\lambda} = \{\boldsymbol{\lambda}_1, \ldots, \boldsymbol{\lambda}_{|\boldsymbol{\lambda}|}\} = 
%    \{
%    \{\lambda_{1, 1}, \ldots, \lambda_{1, |\boldsymbol{\lambda}_1|} \},
%    \ldots,
%    \{\lambda_{|\boldsymbol{\lambda}|, 1}, \ldots, \lambda_{|\boldsymbol{\lambda}|, |\boldsymbol{\lambda}_{|\boldsymbol{\lambda}|}|} \}
%    \}$
%    \item \textsf{Neurological Tests}: $\boldsymbol{\nu} = \{\boldsymbol{\nu}_1, \ldots, \boldsymbol{\nu}_{|\boldsymbol{\nu}|}\} = 
%    \{
%    \{\nu_{1, 1}, \ldots, \nu_{1, |\boldsymbol{\nu}_1|} \},
%    \ldots,
%    \{\nu_{|\boldsymbol{\nu}|, 1}, \ldots, \nu_{|\boldsymbol{\nu}|, |\boldsymbol{\nu}_{|\boldsymbol{\nu}|}|} \}
%    \}$
%    \item \textsf{Medications}: $\boldsymbol{\mu} = \{\boldsymbol{\mu}_1, \ldots, \boldsymbol{\mu}_{|\boldsymbol{\mu}|}\} = 
%    \{
%    \{\mu_{1, 1}, \ldots, \mu_{1, |\boldsymbol{\mu}_1|} \},
%    \ldots,
%    \{\mu_{|\boldsymbol{\mu}|, 1}, \ldots, \mu_{|\boldsymbol{\mu}|, |\boldsymbol{\mu}_{|\boldsymbol{\mu}|}|} \}
%    \}$
%\end{itemize}

\begin{table}[ht]
    \centering
     \begin{tabular}{|c c|} 
     \hline
     Data Components & Permissible Values \\ [0.5ex] 
     \hline
     \age{}         & integer (years) \\
     \sex{}         & 0,1 \\
     \death{}       & 0,1 \\
     \beds{}        & FLR, ICU, OR, TCU, OB, ED  \\ 
     \adm{}         & ICD9 \\
     \dis{}         & ICD9 \\
     \labs{}        & ICD9 \\
     \neuro{}       & (test, result) \\  
     \meds{}        & therapy class \\
     \hline
     \end{tabular}
     \caption{Permissible Values}
     \label{tbl:permvals}
\end{table}

\section{Model Structure} \label{sec:methods}

\begin{figure}
        \includegraphics[width=14cm]{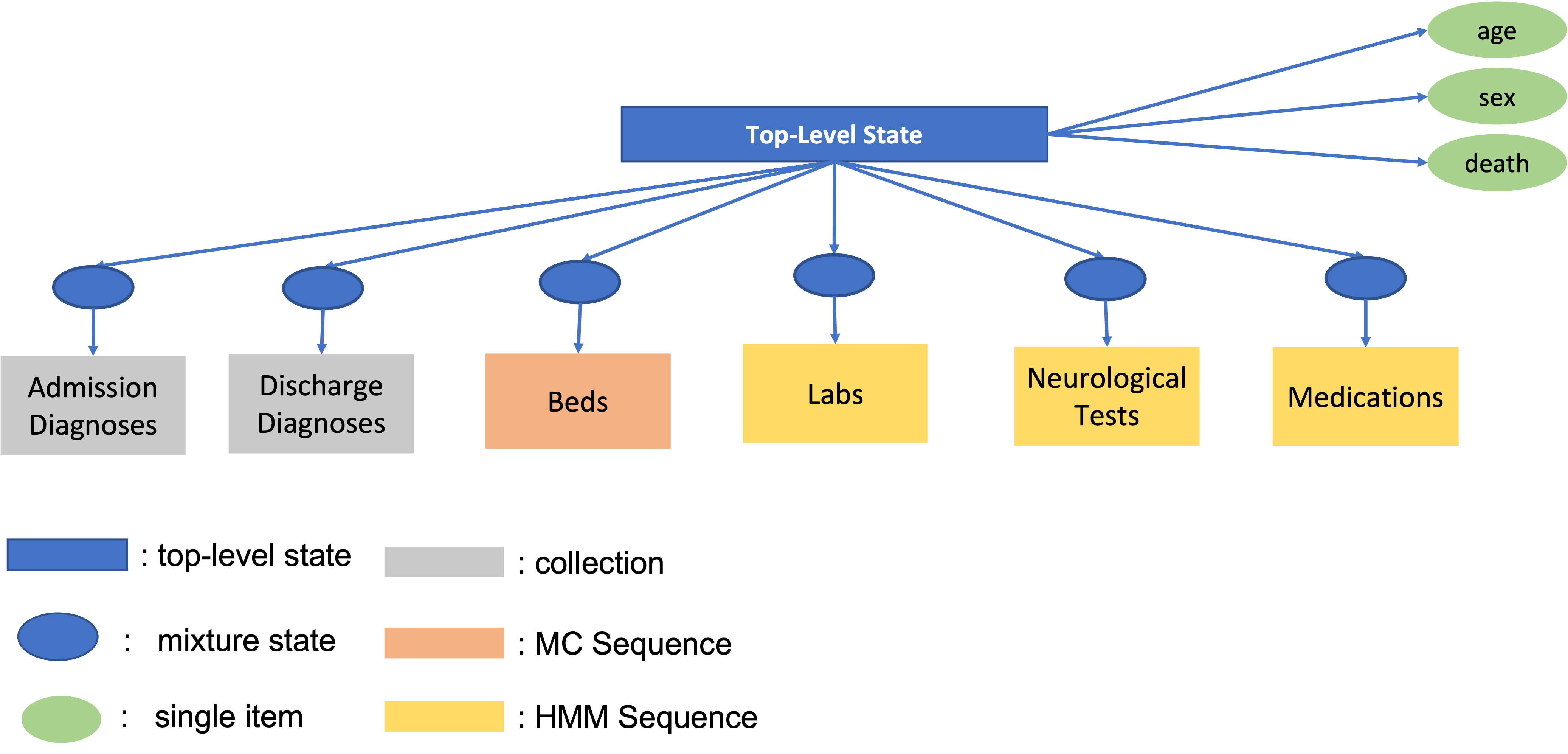}
        \centering
        \caption{Structure of the model. Arrows indicate dependencies in the model. Blue shapes are latent variables and colors correspond to specific model structure types.}
        \label{fig:model}
\end{figure}

In this section we describe the probabilistic model that computes the likelihood of an episode, $f\left(\boldsymbol{y}\right)$.
%Our model is generative and works with irregularly spaced sequences and their combination with scalar elements.
Our model for this collection of data keeps intact its structure and does not require any reformatting or time binning.
Figure \ref{fig:model} shows the structure of the model.
%Colored circles represents single values from the data elements.
Blue shapes represent latent variables and arrows indicate conditional dependencies.
Each leaf in this figure is a sub-model for that specific data sequence (further illustrated in Figure \ref{fig:submodels}).

%Conditioned on this state, each data component uses a specific model:
%\begin{itemize}
%    \item \textsf{Age}: Single quantized Gaussian distribution
%    \item \textsf{Sex}: Single Bernoulli distribution
%    \item \textsf{Death}: Single Bernoulli distribution
%    \item \textsf{Admission Diagnoses}: Mixture of 
%\end{itemize}

%Underneath this state are the three singleton variables: \textsf{Age}, \textsf{Sex}, \textsf{Death}.
%The other multidimensional components are modeled using mixture models conditioned on the top-level state.

%The model uses a number of latent states, one high level state that controls associations between data elements, and one for each of the \textsf{Admission Diagnoses}, \textsf{Discharge Diagnoses}, \textsf{Laboratory Tests}, \textsf{Neurological Tests}, and \textsf{Medications}.

%[diag, labs, neuro, and meds states]

%As described in these steps, all of the data element distributions, including sequence distributions, are dependent on a latent state $z$.
%This latent state characterizes associations between data elements, and is drawn from a random variable $Z$.

The model consists of a top-layer latent state ($Z$) that links the different data elements and sequences.
Under the model, the variables and sequences of an episode are conditionally independent given the latent variable, so that
    %$$
    %    f\left(\boldsymbol{y} | Z=z \right) = f\left(\phi_a | Z=z\right)f\left(\phi_s | %Z=z\right) \cdots f\left(\boldsymbol{\mu} | Z=z\right),
    %$$
    \begin{equation*}
    \begin{split}
                f\left(\boldsymbol{y} | Z=z \right) = & f\left(\phi_a | Z=z\right)f\left(\phi_s | Z=z\right) f\left(\boldsymbol{\beta} | Z=z\right) f\left(\boldsymbol{\alpha} | Z=z\right) \\ & f\left(\boldsymbol{\delta} | Z=z\right) f\left(\boldsymbol{\lambda} | Z=z\right) f\left(\boldsymbol{\nu} | Z=z\right) f\left(\boldsymbol{\mu} | Z=z\right).
    \end{split}
    \end{equation*}
Each term in this expression is a distribution with learned parameters used to model individual data streams.
The distribution of the overall model is formed by summing over the latent variable,
    $$
        f\left(\boldsymbol{y}\right) = \sum_z \alpha_z f\left(\boldsymbol{y} | Z=z \right),
    $$
where $\alpha_z \ge 0$ is a mixing coefficient such that $\sum_z \alpha_z = 1$.
Note that since the latent variable is unknown, the components of the episode are not independent under the model.
%For example, the likelihood of \age{} given \sex{} under this model is $f\left(\phi_a | \phi_s\right) = \frac{\sum_z p_z f\left(\phi_a | Z = z \right)f\left(\phi_s | Z = z \right)}{\sum_z p_z f\left(\phi_s | Z = z \right)} \propto \sum_z p_z f\left(\phi_a | Z = z \right)f\left(Z = z | \phi_s \right) \neq f\left(\phi_a\right)$.
For example, the likelihood of \age{} given \sex{} under this model is $f\left(\phi_a | \phi_s\right) = \frac{\sum_z p_z f\left(\phi_a | Z = z \right)f\left(\phi_s | Z = z \right)}{\sum_z p_z f\left(\phi_s | Z = z \right)} \propto \sum_z p_z f\left(\phi_a | Z = z \right)f\left(Z = z | \phi_s \right)$.
On the other hand, if the model treated all components to be independent, then this would evaluate to the marginal, $f\left(\phi_a\right)$, without considering \textsf{Sex}.

An important aspect of the model is the use of mixture models for the constituent sub-models.
This leads to a layered model structure, resulting in a mixture of mixture models.
Each sub-model has mixing coefficients dependent on the top-layer component.
%, which has several advantages over a single layer model.
%The added structure affords sharing of parameters and lower overall model complexity compared to the single layer version.
The model can be interpreted as building up a representation of the multi-sequence data, starting with states for individual sequences at the lower layer.
The top layer states captures dependencies between these lower layer states.
%This is accomplished by determining the likelihood of co-occurrence between the lower layer states associated with the different sequences.
%The model can be interpreted in a layered fashion, capturing different sequences at the lower-layer, and  that breaks down across data types.
%Each lower-level model consists of states corresponding to each sequence.
%For example, the mixture model for \beds{} describes commonly occurring \beds{ } sequences.
%At the top-layer of the model, the co-occurrence between states is modeled across all of the sequences.
%In this way, we can determine 
%For example, the mixture models for \textsf{Beds} and \textsf{Laboratory Results} capture sets of typical sequences for these data streams, and the higher-level model structure captures co-occurrence between these typical sequences.

Figure \ref{fig:submodels} shows the sub-model structures used to model the individual data sequences.
The colors in this figure correspond to the colors used in Figure \ref{fig:model}.
Each green circle represents a single data element (e.g., one laboratory result or medication).
The sequences can be of differing lengths.
In addition, for \labs{}, \neuro{}, and \meds{}, the number of items for each timepoint can also vary.
The model captures the length of these sequences, as depicted by the \# symbol.
In this way the model can be trained with sequences of differing and arbitrary lengths.
%The \# symbol captures the length of the sequence.
See Appendix A for the generative process describing how episodes can be sampled using the model and Appendix B for a full description of its probability density function.

\subsection*{Collection Model Structure}

\adm{} contain a set of ICD9 codes given at hospital admission, and \dis{} contain a set of ICD9 codes given at discharge.
%\adm{} and \dis{} are sets of ICD9 codes occurring simultaneously (i.e. at hospitalization admission and discharge).
For each of these data elements there is no sequence or temporal information, as they occur simultaneously.
A mixture is used to capture these sets of codes, where each state contains a probability distribution over sets of codes (see Figure \ref{fig:submodels}a).
The number of items (for each state) is described by a Poisson distribution.
Within each state, the likelihood of the codes themselves are given by a categorical distribution over the ICD9 codes.
%Each state is defined by a Poisson distribution and one categorical distribution over all possible diagnoses codes.

\subsection*{Sequence Model Structures}

The \beds{}, \labs{}, \neuro{}, and \meds{} sequences are temporal and have items occurring throughout the hospital stay.
For these, we use model structures that distinguish the different collection times, rather than lumping them into one collection.
Two distinctions between \beds{} and the other sequential streams motivate using different sequence model structures.
First, only one instance of \beds{} can occur at a time.
And secondly, there are a small number of possible \beds{}, whereas there are a much larger set of possible items withing the \labs{}, \neuro{}, and \meds{} sequences.

\begin{figure}
     \centering
     \begin{subfigure}[b]{0.3\textwidth}
         \centering
         \includegraphics[width=\textwidth]{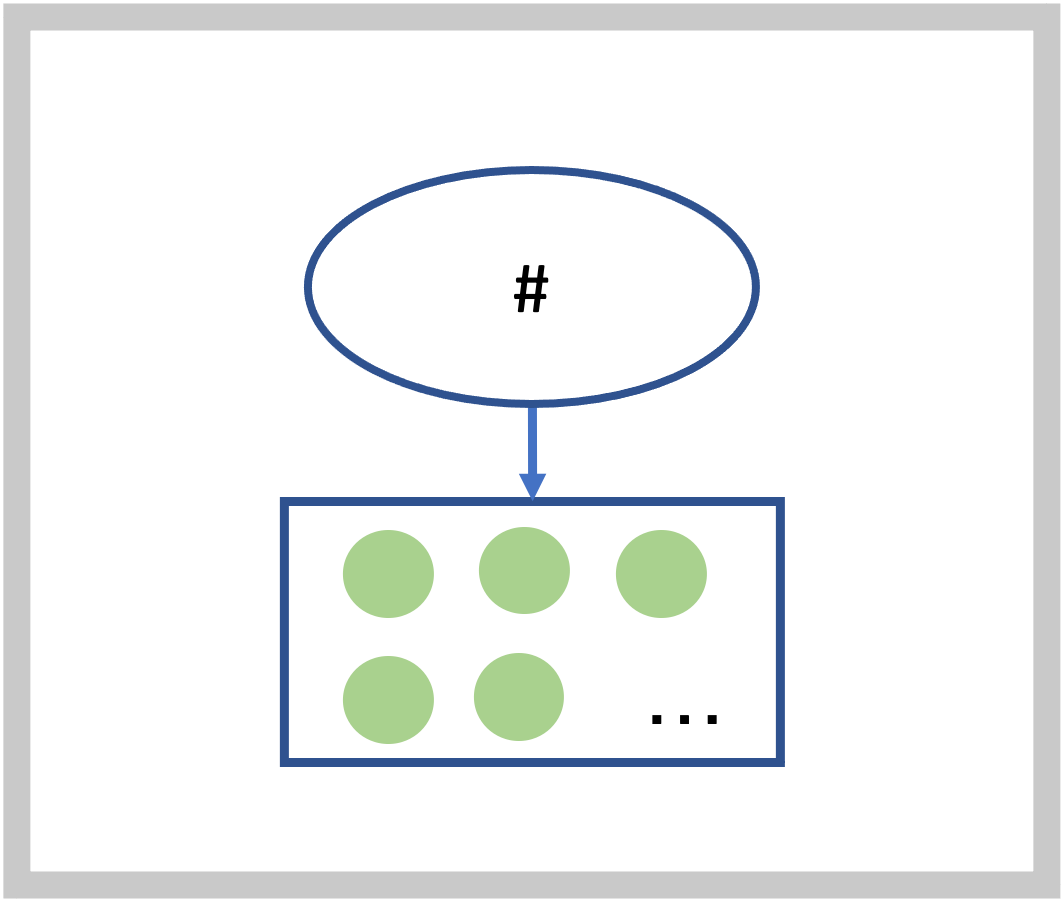}
         \caption{Collection Model}
         \label{fig:y equals x}
     \end{subfigure}
     \hfill
     \begin{subfigure}[b]{0.3\textwidth}
         \centering
         \includegraphics[width=\textwidth]{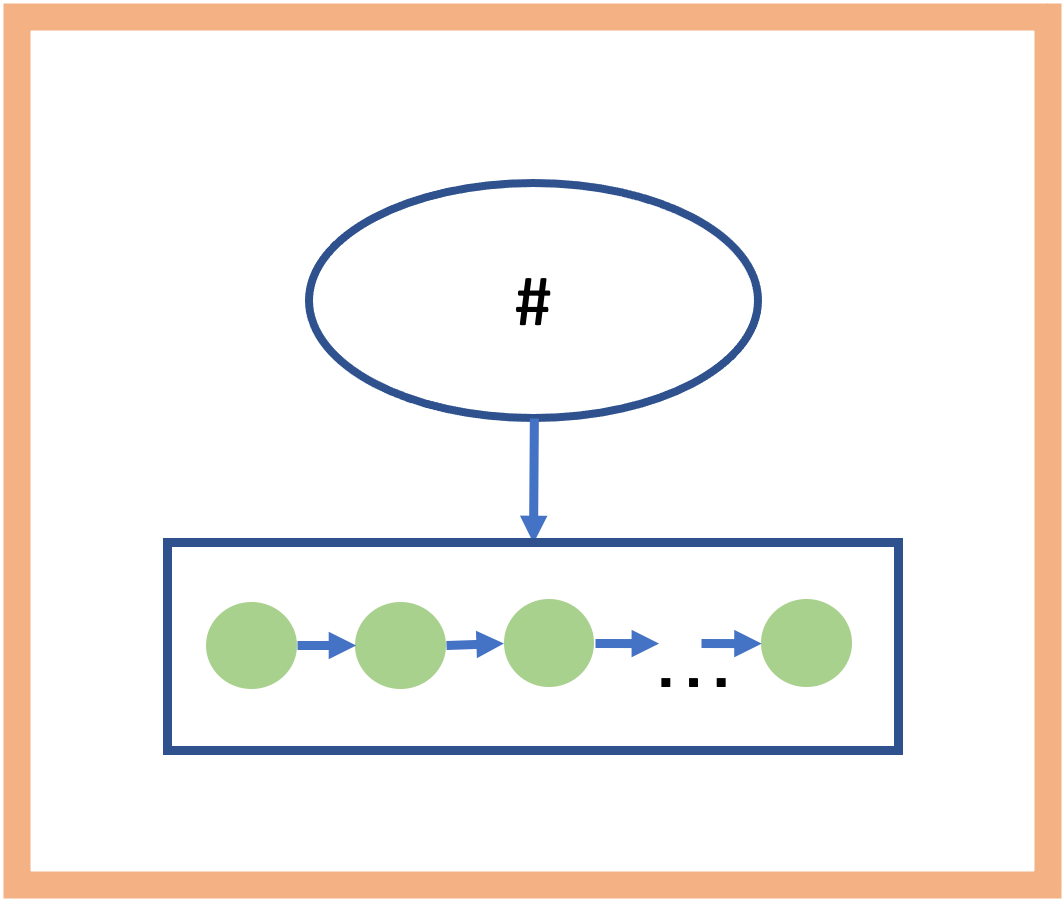}
         \caption{MC Sequence Model}
         \label{fig:three sin x}
     \end{subfigure}
     \hfill
     \begin{subfigure}[b]{0.3\textwidth}
         \centering
         \includegraphics[width=\textwidth]{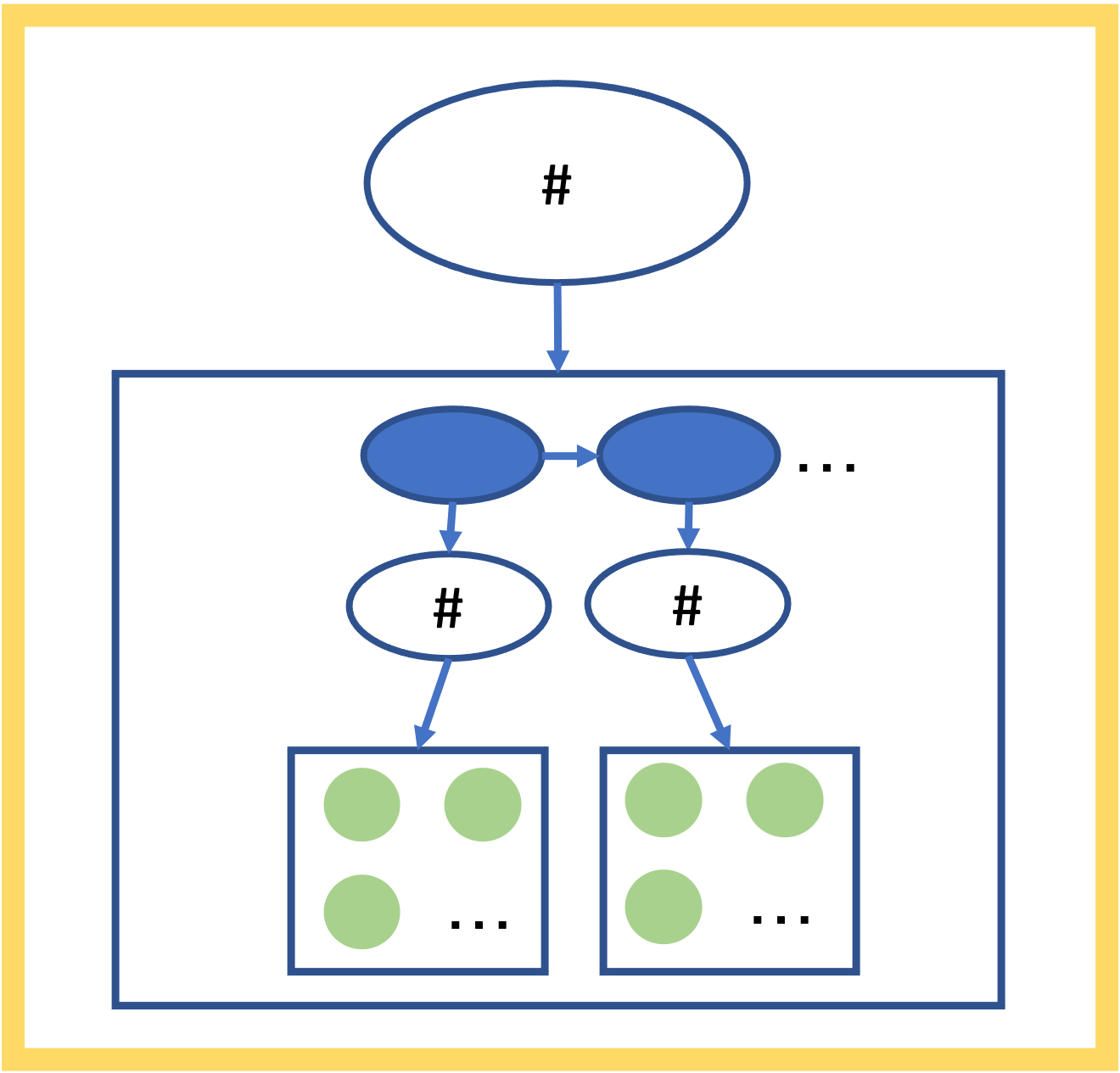}
         \caption{HMM Sequence Model}
         \label{fig:five over x}
     \end{subfigure}
        \caption{Three model structures used for the data streams. \textsf{Admission Diagnoses}, \textsf{Discharge Diagnoses} use the collection model (a), \textsf{Beds} use the MC sequence model (b), and \textsf{Laboratory Results}, \textsf{Medications}, and \textsf{Neurological Tests} use the HMM sequence model (c). Colors correspond to those in Figure \ref{fig:model}. The \# symbols represents the length of the sequence and are captured by Poisson distributions.}
        \label{fig:submodels}
\end{figure}

The model structure used to describe the \beds{} sequence is a mixture of Markov chains, where each state contains a Markov chain (see Figure \ref{fig:submodels}b).
The length of the sequence is characterized by a Poisson distribution.
The Markov chain includes an initial distribution over the set of possible values describing the first item and transition probabilities between sequential items.

For \labs{}, \neuro{}, and \meds{}, we use a mixture of Hidden Markov Models (HMMs) (see Figure \ref{fig:submodels}c).
Similarly to the \beds{} model, a Poisson distribution describes the length of the sequence.
Within each state of the HMM, multiple items can occur simultaneously.
Therefore, each state contains its own length distribution governed by a Poisson rate, and a categorical distribution to describe the collection of items.

Parameter sharing is used to reduce complexity and improve interpretation of the trained model: parameters within the HMM states are shared across mixture states.
This leads to an interpretation of a fixed vocabulary that describes each multi-item instance of \labs{}, \neuro{}, and \meds{}.
The initial and transition probabilities are different across sub-model states to capture a range of different likely sequences.

\subsection*{Top-Layer Latent State}

%The top-layer latent state, $Z$, controls the proportions of the underlying mixture models.
The top-layer latent state selects the subgroup for a patient.
Given data, we can infer this variable to compute probabilities of subgroup membership.
In addition, mixing coefficients of the underlying mixture models are dependent on the top-layer state.
This is a construction of the model.
%These are parameters that impose this structure in the model.
For example, the mixing parameters for the \beds{} sub-model are $p_(\beta,z)$, which is the conditional probability of the \beds{} state given the top-layer latent variable: $\Pr\left(z_\beta|z\right)=p_{\beta,z}$
The same structure is used for all of the sequences.
See Appendix A for definitions for the probability distributions and Table \ref{tbl:params} for notation of the parameters.
%%The goal is to capture co-occurring data elements through their mixture distribution.
%For example, latent state $Z=4$ may describe a higher likelihood for sub-model state \#4 for the \textsf{Laboratory Tests} and sub-model state \#8 for the \textsf{Medications}, etc.
%In this way, the model captures co-occurrence of the data streams.
%This can be used, for example, to examine which \textsf{Beds} are likely to occur with a specific \textsf{Medications} sequence.

Each of the sub-models have common parameters across the top-level components.
For example, the Markov chain parameters for \beds{ state} \#2 is the same for all top-layer states.
The only parameters that change in this example are the mixing coefficients for the \textsf{Beds} mixture.
See Appendices A and B for complete details of the generative model and probability density function.
Table \ref{tbl:ontology} summarizes the three types of latent variables included in the model.

\begin{table}[t]
    \centering
     \begin{tabular}{|c c c c|} 
     \hline
     Layer  & Name              & Notation  & Data Elements \\
     \hline
     1      & Top-layer states  & $z$       & All           \\
     2      & Sub-model states  & $z_\beta, z_\alpha, z_\delta, z_\lambda, z_\nu, z_\mu$       & All           \\
     3      & HMM states        & $\boldsymbol{s}_\lambda, \boldsymbol{s}_\nu, \boldsymbol{s}_\mu$       & \labs{}, \neuro{}, \meds{} \\
     \hline
     \end{tabular}
     \caption{Ontology of Latent States. Each layer contains some number of states in the next lower level.}
     \label{tbl:ontology}
\end{table}

\section{Estimation}

The estimation of the model parameters follows the standard Expectation Maximization (EM) procedure, see e.g. \cite{Moon1996-kb, McLachlan2004-mh}.
Because the conditional dependence structure is a tree, message passing algorithms can be used to compute the required quantities. 
Given a set of episodes, $\boldsymbol{y_1, \ldots, y_N}$ we seek to maximize the log-likelihood $\sum_{i=1}^N \log \sum_z p_z f\left(\boldsymbol{y_i} | Z=z \right)$.
The complete data consists of the episodes, $\boldsymbol{y_1, \ldots, y_N}$, the latent variables, $Z$, $Z_\alpha$, $Z_\delta$, $Z_\beta$, $Z_\lambda$, $Z_\nu$, $Z_\mu$, and the HMM state sequences $\boldsymbol{S}_\lambda$, $\boldsymbol{S}_\nu$, $\boldsymbol{S}_\mu$.
We refer to all of the latent variables as $\boldsymbol{Z}$.
The complete data log-likelihood is,
    $
        \log f\left(\boldsymbol{y}, \boldsymbol{Z}\right) = 
        \sum_{i=1}^N \sum_z I\left(k_i = z\right) \left(\log p_z + \log f\left(\boldsymbol{y_i}, \boldsymbol{Z} | Z=z \right) \right),
    $
where the indicator $I\left(\cdot\right)$ equals 1 when $\cdot$ is true, and otherwise 0.
The expected complete data log-likelihood is,
    $$
        \log f\left(\boldsymbol{y}, \boldsymbol{Z}\right) = 
        \sum_{i=1}^N \sum_z \gamma_i\left(z\right) \left(\log p_z + \mathbb{E} \log f\left(\boldsymbol{y_i}, \boldsymbol{Z} | Z=z \right) \right),
    $$
where $\gamma_i\left(z\right)=f\left(z|\boldsymbol{y}\right)$.
In turn, this likelihood can be written in terms of expected sub-model complete data log-likelihoods,
    \begin{equation*}
    \begin{split}
        \mathbb{E} \log f\left(\boldsymbol{y_i}, \boldsymbol{Z} | Z=z \right)  & = 
        \sum_{z_\alpha} \gamma_{\alpha,i}\left(z_\alpha\right)
        \log f\left(\boldsymbol{\alpha}|Z_\alpha=z_\alpha\right) \\
        & + 
        \sum_{z_\delta} \gamma_{\delta,i}\left(z_\delta\right)
        \log f\left(\boldsymbol{\delta}|Z_\delta=z_\delta\right) \\
        & + 
        \sum_{z_\beta} \gamma_{\beta,i}\left(z_\beta\right)
        \log f\left(\boldsymbol{\beta}|Z_\beta=z_\beta\right) \\
        & + 
        \sum_{z_\lambda} \gamma_{\lambda,i}\left(z_\lambda\right)
        \mathbb{E} \log f\left(\boldsymbol{\lambda}, \boldsymbol{S}_\lambda|Z_\lambda=z_\lambda\right) \\
        & + 
        \sum_{z_\nu} \gamma_{\nu,i}\left(z_\nu\right)
        \mathbb{E} \log f\left(\boldsymbol{\nu}, \boldsymbol{S}_\nu|Z_\nu=z_\nu\right) \\
        & + 
        \sum_{z_\mu} \gamma_{\mu,i}\left(z_\mu\right)
        \mathbb{E} \log f\left(\boldsymbol{\mu}, \boldsymbol{S}_\mu|Z_\mu=z_\mu\right),
    \end{split}
    \end{equation*}
where $\gamma_{x,i}\left(z_x\right)=f\left(z_x|\boldsymbol{y}\right)$ for model $x$.
The expected complete data log-likelihoods for $\boldsymbol{\lambda}$, $\boldsymbol{\nu}$, and $\boldsymbol{\mu}$ are the same as for HMMs and are omitted in this description.
The EM algorithms proceeds by iterating between calculating gamma and maximizing the expected complete data log-likelihood function.

\subsection{Model Selection} \label{sec:training}

In this section, we describe the approach used for model selection.
The hyperparameters of the model are:
\begin{itemize}
    \item the number of top-layer latent states: $|Z|$,
    \item the number of each sub-model latent states: $|Z_\alpha|, |Z_\delta|, |Z_\beta|, |Z_\lambda|, |Z_\nu|, |Z_\mu|$,
    \item the number of HMM states: $|\boldsymbol{S}_\lambda|, |\boldsymbol{S}_\nu|, |\boldsymbol{S}_\mu|$.
\end{itemize}
Note that the number of HMM states is the same across sub-model latent states.
This leads to a total of 10 hyperparameters.
Model selection is performed using the Bayesian Information Criterion (BIC) as a guide.
The BIC penalizes the model fit by a function of the number of parameters: $BIC\left(d\right) = d\ln N - 2\ln f\left(\boldsymbol{y}\right)$, where $d$ is the total parameter count and $N$ is the number of episodes.
To aid in selecting hyperparameters, we first determine those relating to the lower layers, followed by the top-layer.
%Hyperparameters become fixed as we move to higher levels of the model.

To determine these hyperparameters, we first perform linear searches for each sub-model individually.
For the mixture of HMM sequence models (\labs{}, \neuro{}, and \meds{}), we employ the following strategy:
\begin{itemize}
    \item Train a sequence of HMMs with increasing state space,
    \item Compute the BIC for each model, and select the number of HMM states with the lowest corresponding BIC value,
    \item Train a sequence of mixtures of HMMs with increasing state space using the number of HMM states determined previously,
    \item Compute the BIC for each model, and select the number of mixture components with the lowest corresponding BIC value.
\end{itemize}

For the \beds{} model we perform the following:
\begin{itemize}
    \item Train a sequence of mixtures of Markov chains with increasing state space,
    \item Compute the BIC for each model, and select the number of mixture components with the lowest corresponding BIC value.
\end{itemize}

The collection models (\textsf{Admission Diagnoses} and \textsf{Discharge Diagnoses}) use a similar procedure:
\begin{itemize}
    \item Train a sequence of collection mixtures with increasing state space,
    \item Compute the BIC for each model, and select the number of mixture components with the lowest corresponding BIC value.
\end{itemize}

Once these hyperparameters are determined, we can use them to train the entire model.
The number of top-level states can then either be determined again using a linear search over BIC values, or assigned to a pre-selected value.
When training the full model, we do not use model parameters determined previously and re-train the entire model, fixing only the hyperparameters.

%The two hyperparameters, $|Z|$ and $\mathcal{C}_S$ were selected by performing a linear search and computing the BIC on the training set as described in Section \ref{sec:methods}.
%Figure \ref{fig:modelsel} shows the BIC values that were computed on this grid consisting of $|Z|$ taking values 1, 10, 20, 30, 40, 50, 60, 70, 80, 90, 100; and $\mathcal{C}_S$ taking values 1, 2, 5, 10, 15.
%This resulted in 55 trained models.
%From the Figure, we can see that the model order $|Z|=30$ attains a minimum value.
%The $\mathcal{C}_S$ values of 5, 10, and 15 are similar for this value of $|Z|$.
%To err on the side of parsimony, we selected the values: $|Z|=30, \mathcal{C}_S = 5$.

%    \begin{figure}[ht]
%        \centering
%        \begin{subfigure}[b]{0.45\textwidth}
%            \includegraphics[width=\textwidth]{figs/model_bic.png}
%            \caption{BIC values for all trained models}
%        \end{subfigure}
%        \begin{subfigure}[b]{0.45\textwidth}
%            \includegraphics[width=\textwidth]{figs/model_bic_zoom.png}
%            \caption{Zooming in on the local minima}
%        \end{subfigure}
%        \caption{Bayesian Information Criterion (BIC) computed for a series of models with varying hyperparameters. The hyperparameters control the number of latent mixture components and the number of latent states in the Medication sequences.}
%        \label{fig:modelsel}
%    \end{figure}

\section{Results}

\subsection{Training}
We used 195,398 episodes to train the model.
%From our dataset of 244,248 episodes, we randomly selected 80\% for training.
Following the procedure outlined in Section \ref{sec:training}, we determined the number of HMM states for the \labs, \neuro, and \meds.
Then, fixing those hyperparameters, we optimized for the number of mixture states for each sub-model.
Both of these were done by searching over every 10 values, e.g., 10, 20, 30, etc.
Table \ref{tbl:modelsel} shows the optimal number of states found using the model selection approach.
%In this work, we focus on showing details of the trained model.
To enable more straightforward interpretation of the model, we set the number of top-layer states to 10.
We then train all parameters of the final model using these hyperparameters.

    \begin{table}
    \begin{center}
    \begin{tabular}{ |c|c|c| } 
        \hline
        \bf{Category} & \bf{Sub-model States} & \bf{HMM States} \\
        \hline
        \textsf{Diagnoses} & 10 & N/A \\ 
        \beds{} & 10 & N/A \\ 
        \labs{} & 10 & 90 \\
        \neuro{} & 10 & 70 \\
        \meds{} & 10 & 40 \\
    \hline
    \end{tabular}
    \caption{\label{tbl:modelsel} Number of HMM states (for \labs{}, \neuro{}, and \meds{}) and mixture states determined by evaluating the BIC score on the training data. The \textsf{Diagnoses} states are shared between \adm{} and \dis{}.}
    \end{center}
    \end{table}

\subsection{Top-layer Components with \age{}, \sex{}, and \death{} Probabilities}
\begin{figure}
     \centering
     \begin{subfigure}[t]{0.49\textwidth}
         \centering
         \includegraphics[width=\textwidth]{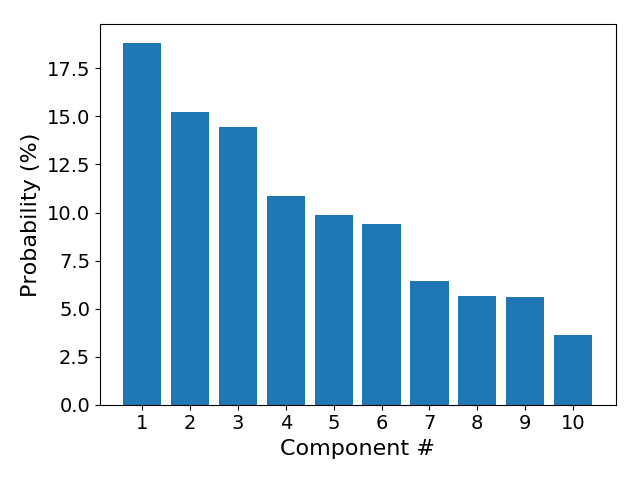}
         \caption{Component Probabilities}
         \label{subfig:compstat_prob}
     \end{subfigure}
     \hfill
     \begin{subfigure}[t]{0.49\textwidth}
         \centering
         \includegraphics[width=\textwidth]{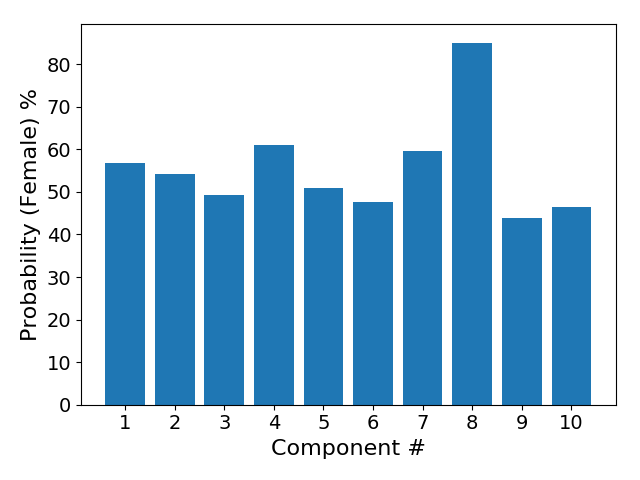}
         \caption{\sex{}}
         \label{subfig:compstat_sex}
     \end{subfigure}
     \hfill
     \begin{subfigure}[t]{0.49\textwidth}
         \centering
         \includegraphics[width=\textwidth]{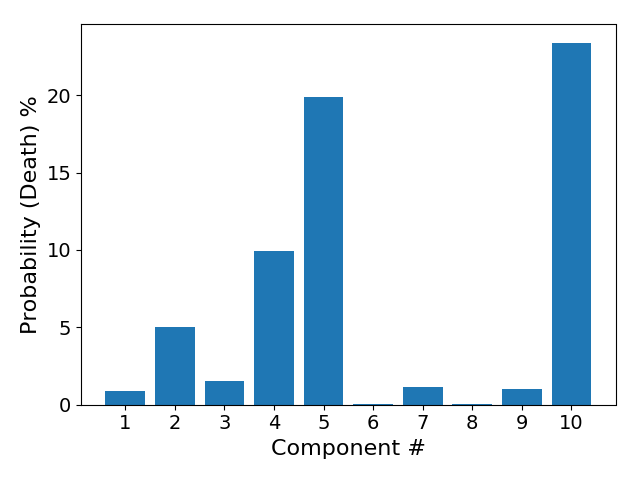}
         \caption{\death{}}
         \label{subfig:compstat_mort}
     \end{subfigure}
     \begin{subfigure}[t]{0.49\textwidth}
         \centering
         \includegraphics[width=\textwidth]{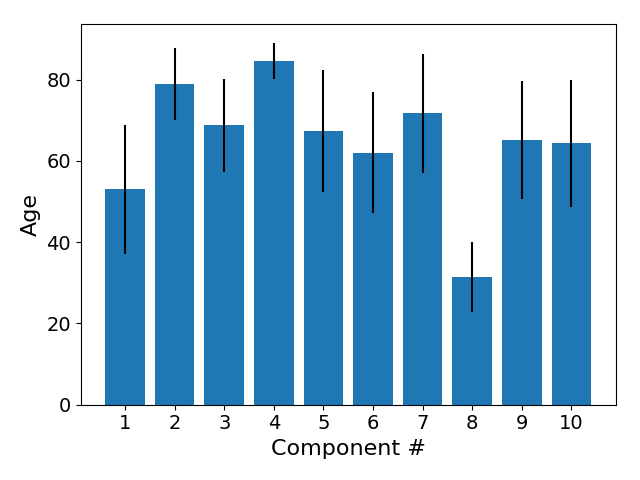}
         \caption{\age{}}
         \label{subfig:compstat_age}
     \end{subfigure}
        \caption{Top-level component probabilities (\ref{subfig:compstat_prob}), probability of  \sex{} (\ref{subfig:compstat_sex}), \death{} (\ref{subfig:compstat_mort}), and \age{} mean and spread (\ref{subfig:compstat_age}) for each learned component. The components are sorted from highest to lowest probability and correspond across charts.}
        \label{fig:compstat}
    \end{figure}

The top-layer component probabilities are shown in Figure \ref{subfig:compstat_prob} in decreasing order.
This ordering of the components is preserved throughout the paper.
Figures \ref{subfig:compstat_sex}, \ref{subfig:compstat_mort}, and \ref{subfig:compstat_age} show learned parameters for scalar variables (\age{}, \sex{}, \death{}) across these 10 components.
Components with the highest probability of \death{} are 4, 5, and 10.
%Figure \ref{subfig:compstat_mort} shows that the components' enrichment of \death{} vary.

\subsection{Enrichment Analysis- Mortality} \label{sec:res_mort}
Figure \ref{subfig:compstat_mort} shows that the probability of \death{} varies across components.
Enrichment of mortality can be defined with respect to each top-layer component.
%%%%%%NOTE subgroup patients and sequence prototypes
Further, we can determine sub-model states (corresponding to e.g. \beds{}, \meds{}, etc.) that are enriched for mortality.
%We use the trained model to identify sequence states that contribute to this enrichment.
This is done by computing
$$
    f\left(\death|z_x\right) \propto \sum_z \alpha_z f\left(z_x|z\right)f\left(\death{}|z\right),
$$
where $z$ is the top-layer state and $z_x$ is the state for sub-model $x$.
This is performed for each sub-model to determine which states have the greatest and least contribution to \death{}.
%Table \ref{tbl:enr_mort} shows the lowest and highest risk states for \death{} across sequences.
Table \ref{tbl:enr_mort} shows the probability of \death{} given the state for each model.

 \begin{table}
    \begin{center}
    \begin{tabular}{ |c|c|c|c|c|c|c|c|c|c|c| } 
        \hline
        \bf{Sub-Model} & 1 & 2 & 3 & 4 & 5 & 6 & 7 & 8 & 9 & 10\\
        \hline
        \hline
        %\hline
        \beds{} & 0.03 & 0.08 & 0.90 & 4.18 & 4.43 & 5.09 & 8.35 & 20.72 & 23.67 & 32.55 \\ 
        \hline
        \hline
        \labs{} & 0.10 & 3.34 & 3.61 & 3.66 & 6.09 & 6.60 & 6.61 & 19.95 & 24.63 & 25.38 \\
        \hline
        \hline
        \neuro{} & 1.22 & 2.00 & 2.18 & 4.46 & 8.20 & 8.37 & 14.20 & 18.33 & 20.34 & 20.70 \\
        \hline
        \hline
        \meds{} & 0.04 & 0.08 & 1.51 & 2.22 & 3.12 & 3.46 & 5.84 & 15.84 & 31.37 & 36.52 \\
        \hline
    \end{tabular}
    \caption{\label{tbl:enr_mort} Probability of \death{} (\%) for each sub-model state for \beds{}, \labs{}, \neuro{} and \meds{}. States are sorted from lowest to highest probability.}
    \end{center}
\end{table}

\begin{figure} 
     \centering
     \begin{subfigure}[t]{0.49\textwidth}
         \centering
         \includegraphics[width=0.8\textwidth]{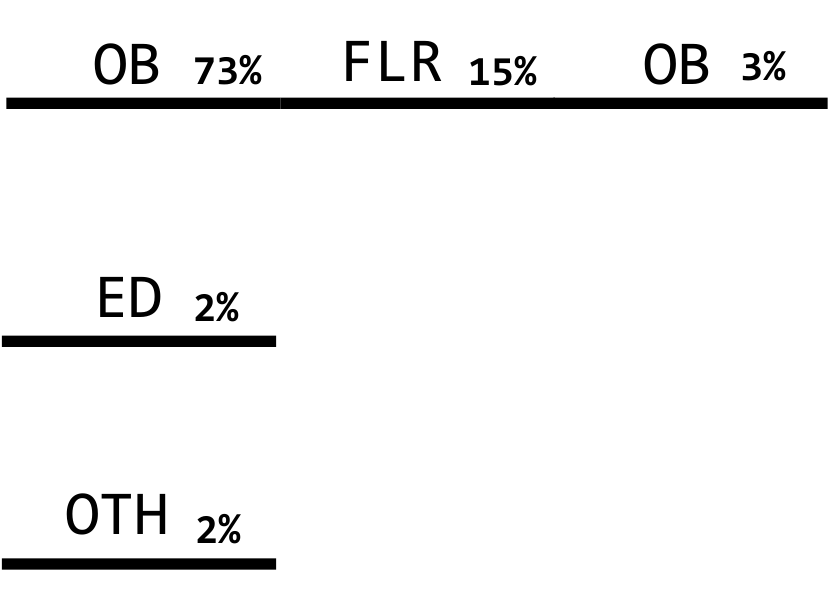}
         \caption{}
         \label{subfig:enr_mort_bed_low}
     \end{subfigure}
     \begin{subfigure}[t]{0.49\textwidth}
         \centering
         \includegraphics[width=0.9\textwidth]{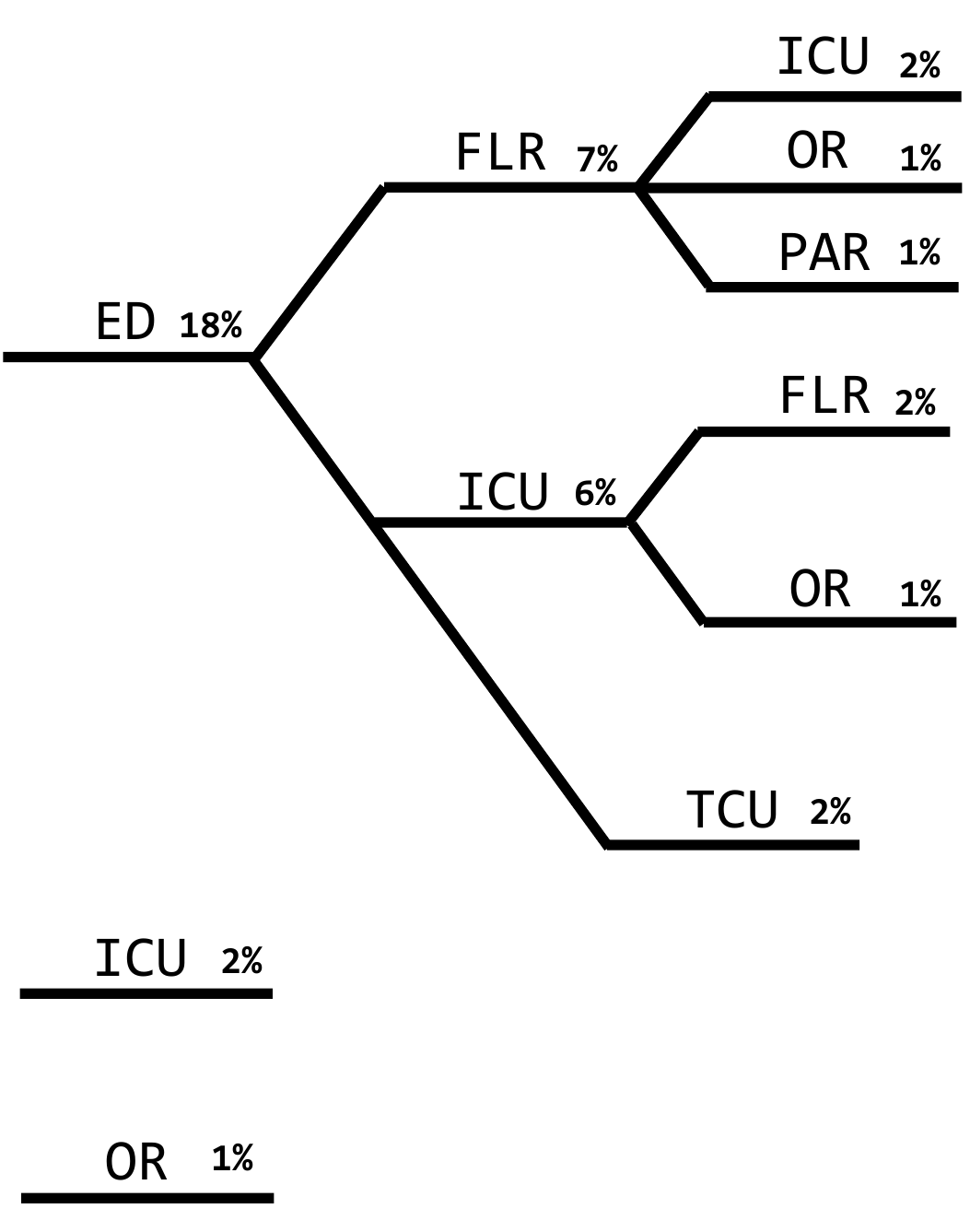}
         \caption{}
         \label{subfig:enr_mort_bed_high}
     \end{subfigure}
     \hfill
     \begin{subfigure}[t]{0.49\textwidth}
         \centering
         \includegraphics[width=\textwidth]{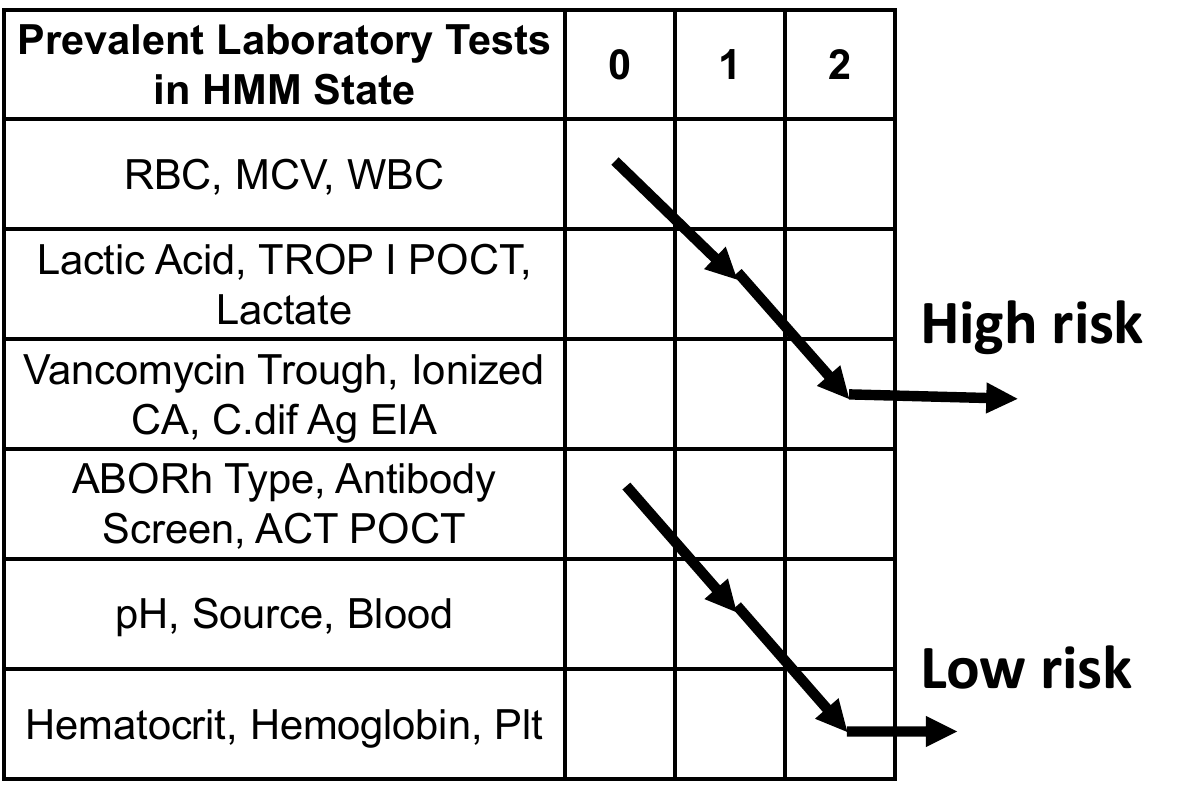}
         \caption{}
         \label{subfig:enr_mort_labs}
     \end{subfigure}
     \hfill
     \begin{subfigure}[t]{0.49\textwidth}
         \centering
         \includegraphics[width=\textwidth]{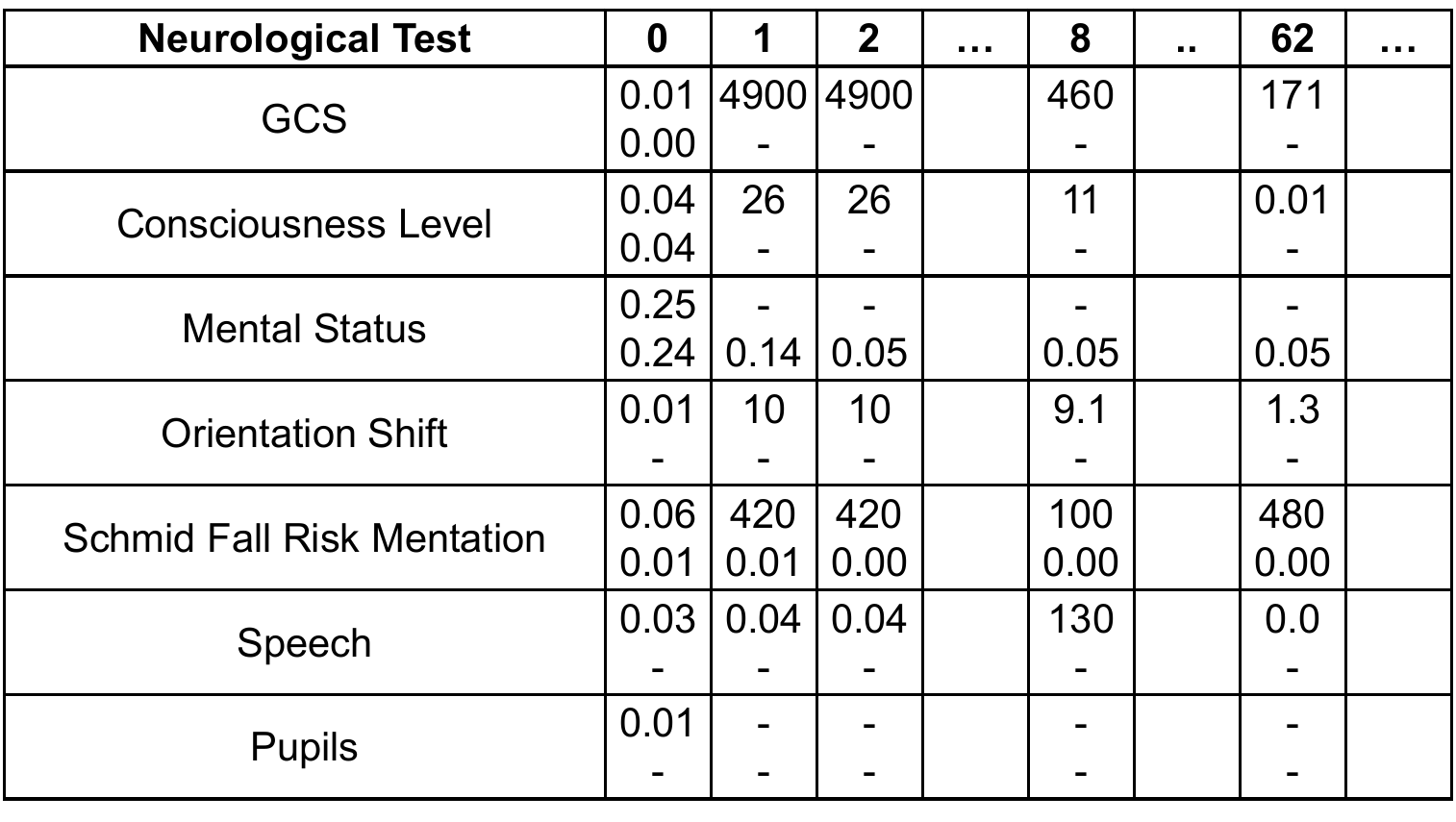}
         \caption{}
         \label{subfig:enr_mort_neuro}
     \end{subfigure}
     \hfill
     %\begin{subfigure}[t]{0.99\textwidth}
     %   \centering
     %   \begin{tabular}{ |c|c|c| } 
     %       \hline
     %       \bf{Prevalent Therapy Classes} & \bf{Timepoint(s)} \\
     %       \hline
     %       Elect/Caloric/H2O, Analgesics & 1 \\ 
     %       \hline
     %      Antibiotics, Analgesics & 2-3 \\
     %       \hline
     %       Anesthetics, Autonomic Drugs & 4-8 \\
     %       \hline
     %       Elect/Caloric/H2O, Other & 9-11 \\
     %       \hline
     %       Psychotherapeutic Drugs & 12- \\
     %   \hline
     %   \end{tabular}
     %   \caption{ \meds{}}
     %   \label{subfig:enr_mort_meds}
    %\end{subfigure}
     \begin{subfigure}[t]{0.99\textwidth}
         \centering
         \includegraphics[width=\textwidth]{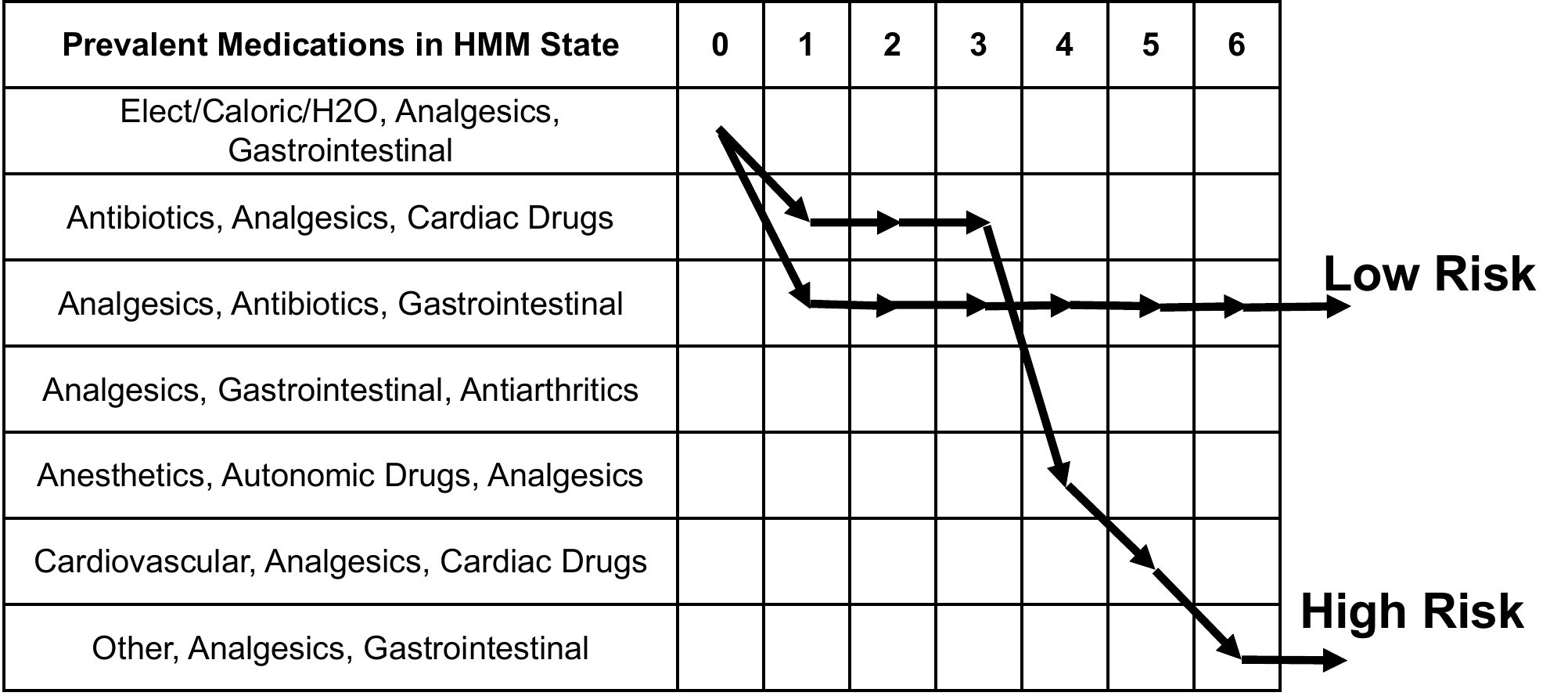}
         \caption{}
         \label{subfig:enr_mort_meds}
     \end{subfigure}
        \caption{Lowest and Highest risk states for \death{} enrichment. Figure \ref{subfig:enr_mort_bed_low} shows the \beds{} sequences with the lowest enrichment. Figure \ref{subfig:enr_mort_bed_high} shows the sequences with the highest enrichment. Figure \ref{subfig:enr_mort_labs} shows the most likely sequences through HMM states for the lowest and highest enrichment. Figure \ref{subfig:enr_mort_neuro} shows likelihood ratios for abnormal vs normal \neuro{} results for the low risk (bottom) and high risk (top) states. Figure \ref{subfig:enr_mort_meds} shows the most likely \meds{} sequences through he HMM states for the two states.}
        \label{fig:enr_mort}
\end{figure}

%We further examine states that contribute to enrichment of \death{}.
The likelihood of \beds{} sequences for the lowest \death{} risk state is shown in Figure \ref{subfig:enr_mort_bed_low}.
These sequences are determined from the Markov chain for that state.
Probabilities in the tree graph indicate the likelihood of \beds{} sequences terminating at the corresponding edge.
For example, the OB $\rightarrow$ FLR sequence has a likelihood of 15\%.
Only sequences with likelihood of 1\% or greater are shown.
Figure \ref{subfig:enr_mort_bed_high} shows the sequence likelihoods for the highest risk state.

Figures \ref{subfig:enr_mort_labs}, \ref{subfig:enr_mort_neuro}, and \ref{subfig:enr_mort_meds} show the most likely progression through the \labs{}, \neuro{}, and \meds{} sequences for the lowest and highest risk states.
Each row corresponds to an HMM state, with the three most prevalent items in the first column.
Each numbered column represents the time step, and the arrows indicate the most likely sequence through the HMM states for the high and low risk HMM states.

In our trained models, after some number of timepoints, the underlying Hidden Markov Model (HMM) enters an HMM state (rows in Figures \ref{subfig:enr_mort_labs}, \ref{subfig:enr_mort_neuro}, and \ref{subfig:enr_mort_meds}) that has itself as the next most likely HMM state.
Thus, for the two risk states considered in the Figure (low and high mortality risk), the models have a terminal HMM state, where it the subject is most likely to remain in.
For this reason, we show the initial sequence the states until this phenomenon occurs, after which the most likely HMM state does not change.

%Each row in Figure \ref{subfig:enr_mort_labs} corresponds to a \labs{} HMM state, with the three most prevalent items in the first column.
%Each numbered column represents the time step, and the arrows indicate the most likely sequence through the HMM states for the high and low risk \labs{} states.
%This is computed by finding the most likely state sequence in the \labs{} HMM for these two states.

For \neuro{}, Table \ref{subfig:enr_mort_neuro} shows the likelihood ratio between Normal and Abnormal results for each \neuro{} test.
Within each cell, the top value is the ratio for the high risk state, and the bottom value is the ratio for the low risk state.
Entries with `$-$' indicate that the test is unlikely to be administered (probability $<$ 1\%).
For example, initially (at timestep 0), Consciousness Level is approximately 25 (1/0.04) times as likely to be normal than abnormal for both the high and low risk groups.
At timestep 1, however, abnormal results are 26 times as likely for the high risk state, while the test is unlikely to be administered for the low risk state.

%Figure \ref{subfig:enr_mort_meds} shows the most likely \meds{} sequences for the high and low risk states over time.
%Each row corresponds to an HMM state, and the first column contains the three most likely therapy classes for that HMM state.

%\subsection{Enrichment Analysis- Sepsis}
%
%To determine the risk level for Sepsis, we can compute the likelihood of a Sepsis discharge diagnosis code (995.91) for each state.
%This is computed by,
%$$
%    e_s \propto \sum_z \sum_d \alpha_z f\left(s|z\right)f\left(d|z\right)f\left(\text{Sepsis}|d\right),
%$$
%where $z$ is the top-layer component, $s$ is the state, and $d$ is the \dis{} state.
%Table \ref{tbl:enr_sepsis} shows the highest and lowest risk state for each of the sequence models.
%
%The separation between high and low risk is much smaller than in the case of \death{} enrichment (Table \ref{tbl:enr_mort}).
%
% \begin{table}
%    \begin{center}
%    \begin{tabular}{ |c|c|c| } 
%        \hline
%        \bf{Sequence} & \bf{Lowest Risk State} & \bf{Highest Risk State} \\
%        \hline
%        \beds{} & D (7.1) & F (11.6) \\ 
%        \hline
%        \labs{} & F (5.3) & B (11.8) \\
%        \hline
%        \neuro{} & G (8.1) & J (11.1) \\
%        \hline
%        \meds{} & C (8.0) & F (11.5) \\
%    \hline
%    \end{tabular}
%    \caption{\label{tbl:enr_sepsis} Lowest and highest risk for Sepsis discharge diagnosis code (995.91) among the sequences according to the model. The probability of Sepsis given that the state is true is given in parenthesis.}
%    \end{center}
%\end{table}

%\paragraph{State compositions}
\subsection{Sub-model State Distributions}
\begin{figure} 
     \centering
     \begin{subfigure}[t]{0.49\textwidth}
         \centering
         \includegraphics[width=\textwidth]{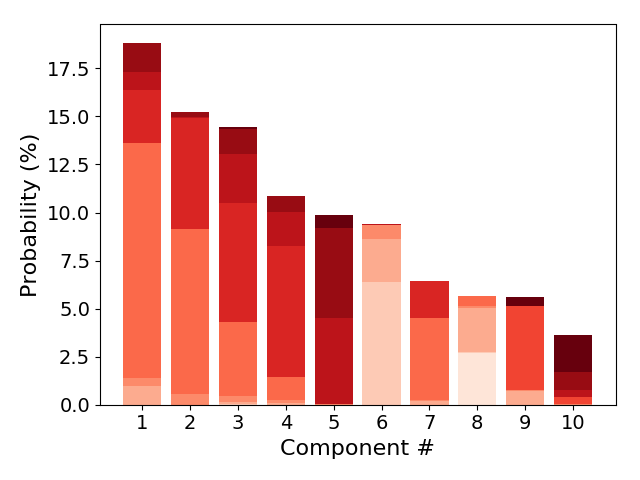}
         \caption{\beds{}}
         \label{subfig:compdist_bed}
     \end{subfigure}
     \hfill
     \begin{subfigure}[t]{0.49\textwidth}
         \centering
         \includegraphics[width=\textwidth]{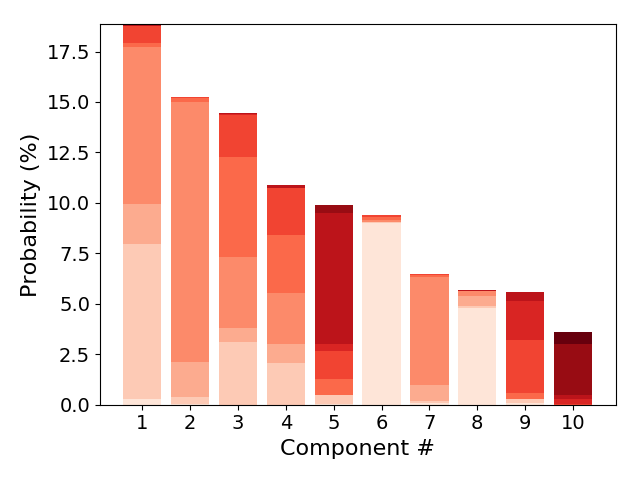}
         \caption{\labs{}}
         \label{subfig:compdist_labs}
     \end{subfigure}
     \hfill
     \begin{subfigure}[t]{0.49\textwidth}
         \centering
         \includegraphics[width=\textwidth]{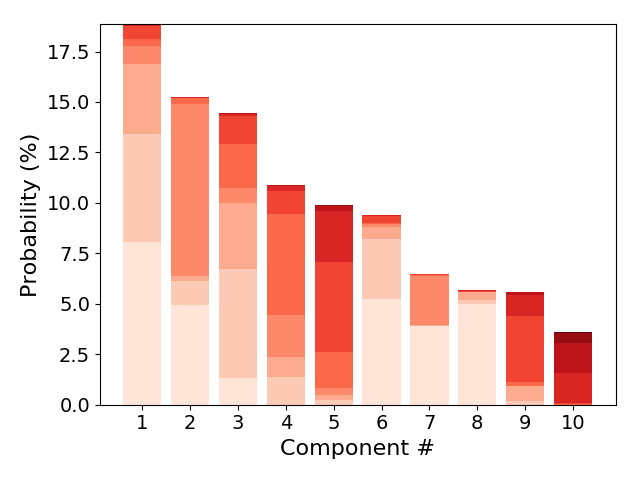}
         \caption{\neuro{}}
         \label{subfig:compdist_neuro}
     \end{subfigure}
     \begin{subfigure}[t]{0.49\textwidth}
         \centering
         \includegraphics[width=\textwidth]{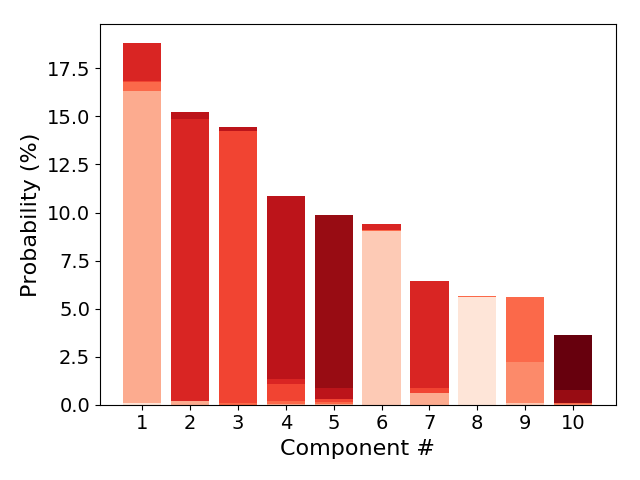}
         \caption{\meds{}}
         \label{subfig:compdist_meds}
     \end{subfigure}
        \caption{State distributions across the component states. Each component has a distribution across the 10 underlying states.}
    \label{fig:compdist}
\end{figure}

\begin{figure} 
     \centering
     \begin{subfigure}[t]{0.49\textwidth}
         \centering
         \includegraphics[width=\textwidth]{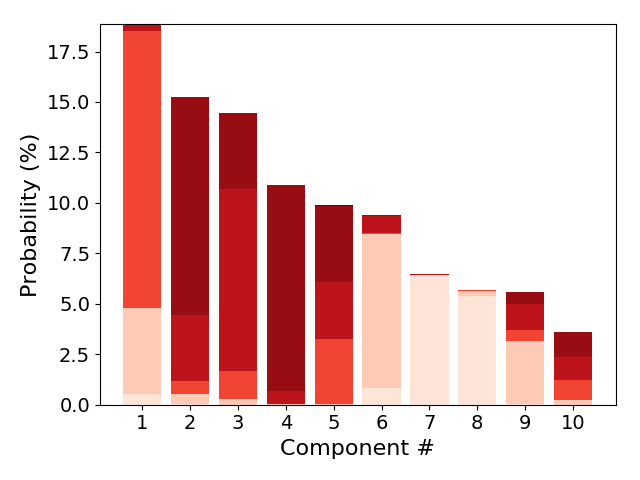}
         \caption{\adm{}}
         \label{subfig:diagdist_adm}
     \end{subfigure}
     \hfill
     \begin{subfigure}[t]{0.49\textwidth}
         \centering
         \includegraphics[width=\textwidth]{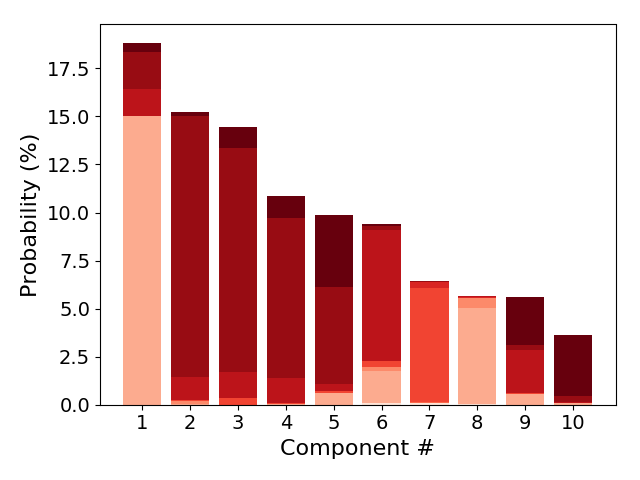}
         \caption{\dis{}}
         \label{subfig:diagdist_disch}
     \end{subfigure}
        \caption{\adm{} and \dis{} state distributions across the component states. Each component has a distribution across the 10 underlying states.}
    \label{fig:diagdist}
\end{figure}

%As described in Section \ref{sec:methods}, sub-models for the different sequences each have a number of states.
Figure \ref{fig:compdist} shows the distributions of the sub-model states for each of the top-layer components.
The height of each bar is the probability $\alpha_z$ of each component (same as Figure \ref{fig:compdist}).
Within each bar are 10 sections, each one corresponding to a sub-model state.
The height of these bars are the probabilities of the sub-model state for the given top-layer state, $\alpha_z f(z_x|z)$.
They are ranked and color-coded by the mortality enrichment probabilities computed in Section \ref{sec:res_mort}, from lowest risk (light red) to highest risk (dark red).
Figure \ref{fig:diagdist} shows the state distributions across these 10 components for \adm{} and \dis{}.
%(Figure \ref{fig:statecode} shows the color associated with each state).

%name of each state (A, B, C, etc.) associated with each color.
%The height of each bar represents the component probabilities $\alpha_z$, and the height of each segment within the bar is the probability of the state $s$ within each component, $\alpha_z f(s|z)$.
%For example, we can see that state A is the most likely \beds{} state for component 1.
%Figure \ref{fig:diagdist} shows the state distributions across these 10 components for \adm{} and \dis{}.

%\begin{figure}
%    \centering
%    \includegraphics[width=\textwidth]{figs/state_key.pdf}
%    \caption{Color key for sequence states. Each state corresponds to a sub-model mixture component.}
%    \label{fig:statecode}
%\end{figure}

\subsection{Sequence lengths}
\begin{figure} 
     \centering
     \begin{subfigure}[t]{0.49\textwidth}
         \centering
         \includegraphics[width=\textwidth]{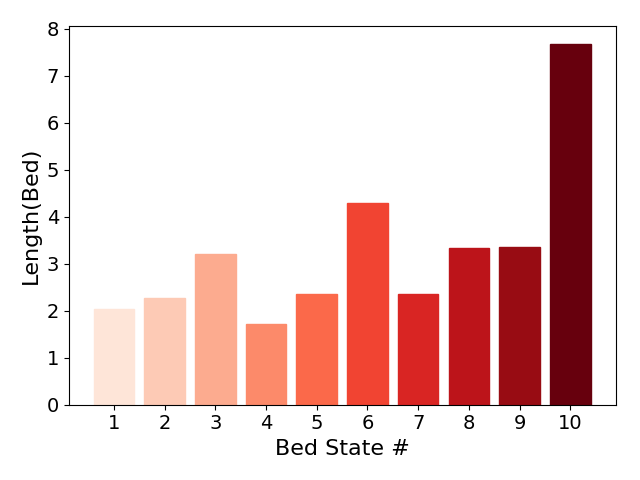}
         \caption{}
         \label{subfig:statelen_bed}
     \end{subfigure}
     \hfill
     \begin{subfigure}[t]{0.49\textwidth}
         \centering
         \includegraphics[width=\textwidth]{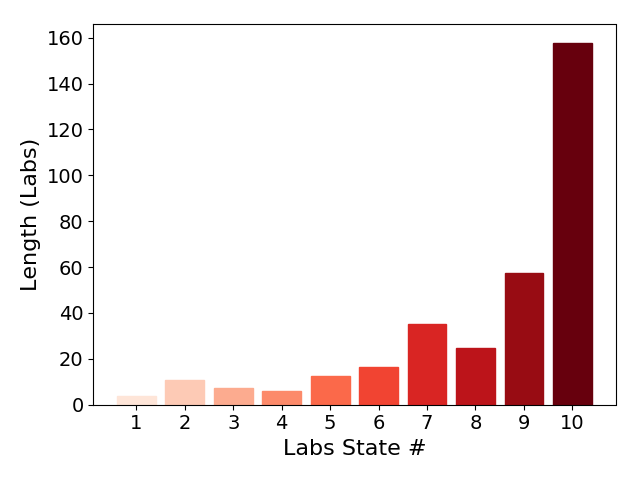}
         \caption{}
         \label{subfig:statelen_labs}
     \end{subfigure}
     \hfill
     \begin{subfigure}[t]{0.49\textwidth}
         \centering
         \includegraphics[width=\textwidth]{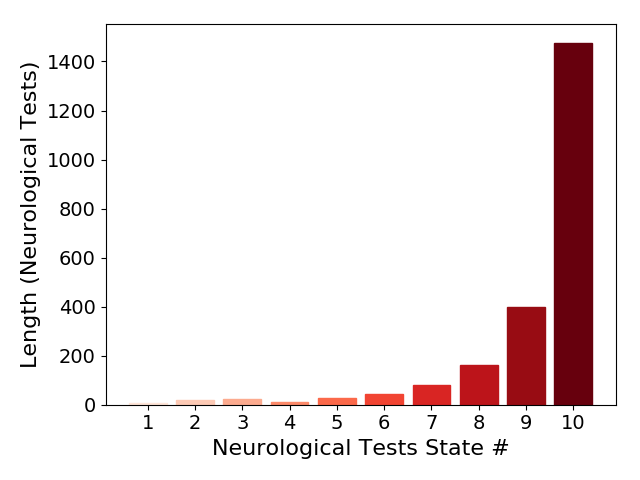}
         \caption{}
         \label{subfig:statelen_neuro}
     \end{subfigure}
     \hfill
     \begin{subfigure}[t]{0.49\textwidth}
         \centering
         \includegraphics[width=\textwidth]{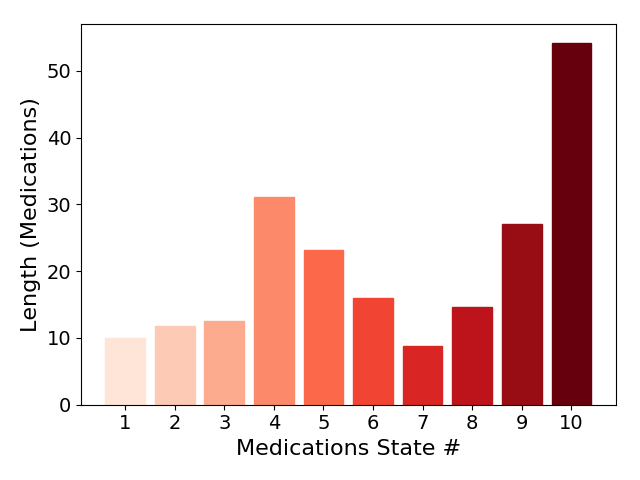}
         \caption{}
         \label{subfig:statelen_meds}
     \end{subfigure}
        \caption{Learned mean length parameters for the 10 states in each of \beds{}, \labs{}, \neuro{}, and \meds{}. The colors correspond to those in Figure \ref{fig:compdist}.}
        \label{fig:statelen}
\end{figure}

%Next we examine the learned states for each data stream.
The length parameters, which are the mean number (Poisson rate) of items in each state for \beds{}, \labs{}, \neuro{}, and \meds{} are shown in Figure \ref{fig:statelen}.
The same color-coding by mortality enrichment is used in these figures as in Figures \ref{fig:compdist} and \ref{fig:diagdist}.
%states in these charts are color-coded and correspond across figures for each sub-model.
%For example, state 4 is red throughout, but these are different states for \beds{} and \labs{}.

\subsection{\adm{} and \dis{}} \label{sec:adm_dis}
Of the 10 states shared across \adm{} and \dis{}, 5 were predominantly associated with \adm{} and 5 with \dis{}.
Table \ref{tbl:admdiagcodes} shows the 5 most likely ICD9 diagnosis codes for each state associated with \adm{}.
Each column corresponds to a state and column headings contain the prevalence of the state, $\sum_z w_{z_s}$.
The states are sorted in ascending order according to their mortality enrichment.
%This was determined purely by estimating the models from the data, and not prespecified.
Similarly, Table \ref{tbl:dischdiagcodes} shows the 5 most likely ICD9 codes for states associated with \dis{}.

\begin{table}[] 
        \centering
        \begin{tabular}{p{6.5cm}|c|c|c|c|c}
           \bf{Diagnosis} \textbackslash \bf{Prevalence} & 13.2\% & 16.3\% & 20.6\% & 19.4\% & 30.5\% \\ 
           \hline
           (648.91) Other current conditions classifiable elsewhere of mother, delivered, with or without mention of antepartum condition          & X & - & - & - & - \\
           (278.00) Obesity                                         & X & X & - & - & - \\
           (V221) Supervision of other normal pregnancy             & X & - & - & - & - \\
           (644.21) Early onset of delivery, delivered, with or without mention of antepartum condition                                                & X & - & - & - & - \\
           (493.90) Asthma                                          & X & - & - & - & - \\
           (401.9) Unspecified essential hypertension               & - & X & X & X & X \\
           (272.4) Other and unspecified hyperlipidemia             & - & X & X & X & X \\
           (530.81) Esophageal reflux                               & - & X & X & - & - \\
           (305.1) Tobacco use disorder                             & - & X & - & - & - \\
           (038.9) Unspecified septicemia                           & - & - & X & - & - \\
           (995.91) Sepsis                                          & - & - & X & - & - \\
           (250.40) Diabetes with renal manifestations              & - & - & - & X & - \\
           (357.2) Polyneuropathy in diabetes                       & - & - & - & X & - \\
           (403.90) Hypertensive chronic kidney disease             & - & - & - & X & X \\
           (427.31) Atrial fibrillation                             & - & - & - & - & X \\
           (428.0) Congestive heart failure                         & - & - & - & - & X \\
        \end{tabular}
        \caption{Most likely (top 5) ICD codes for the states predominantly in \adm{}. Column headings contain the prevalence of the state.}
    \label{tbl:admdiagcodes}
\end{table}

\begin{table}[] 
        \centering
        \begin{tabular}{p{6.5cm}|c|c|c|c|c}
           \bf{Diagnosis} \textbackslash \bf{Prevalence} & 22.9\% & 6.9\% & 14.7\% & 41.5\% & 12.4\% \\
           \hline
            (V15.82) Personal history of tobacco use                & X & X & X & X & X \\
            (789.00) Abdominal pain                                 & X & - & - & - & - \\
            (780.60) Fever                                          & X & - & - & - & - \\
            (V27.0) Outcome of delivery, single liveborn            & X & - & - & - & - \\
            (682.9) Cellulitis and abscess of unspecified sites     & X & - & - & - & - \\
            (272.4) Other and unspecified hyperlipidemia            & - & X & - & - & - \\
            (401.9) Unspecified essential hypertension              & - & X & - & - & - \\
            (530.81) Esophageal reflux                              & - & X & - & - & - \\
            (599.0) Urinary tract infection                         & - & X & - & - & - \\
            (E849.0) Home accidents                                 & - & - & X & - & - \\
            (E849.7) Accidents occurring in residential institution & - & - & X & - & X \\
            (V58.66) Long-term use of aspirin                       & - & - & X & X & X \\
            (E849.9) Accidents occurring in unspecified place       & - & - & X & - & - \\
            (V58.67) Long-term use of insulin                       & - & - & - & X & X \\
            (V58.61) Long-term use of anticoagulants                & - & - & - & X & X \\
            (412) Old myocardial infarction                         & - & - & - & X & - \\
        \end{tabular}
        \caption{Most likely (top 5) ICD codes for the states predominantly in \dis{}. Column headings contain the prevalence of the state.}
    \label{tbl:dischdiagcodes}
\end{table}

\subsection{\beds{} sequences}
\begin{figure}
     \centering
     \begin{subfigure}[t]{0.24\textwidth}
         \centering
         \includegraphics[width=\textwidth]{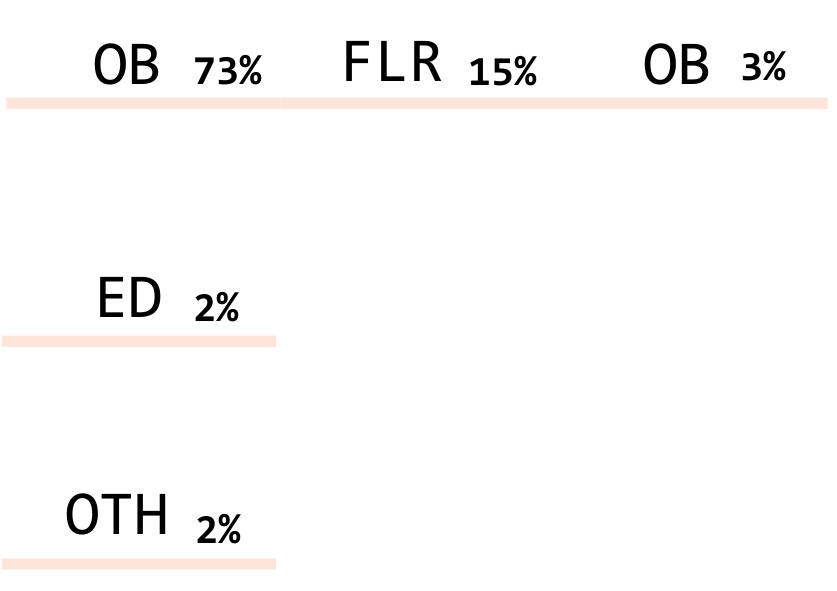}
         \caption{S1}
         \label{subfig:bedseq_1}
     \end{subfigure}
     \hfill
     \begin{subfigure}[t]{0.24\textwidth}
         \centering
         \includegraphics[width=\textwidth]{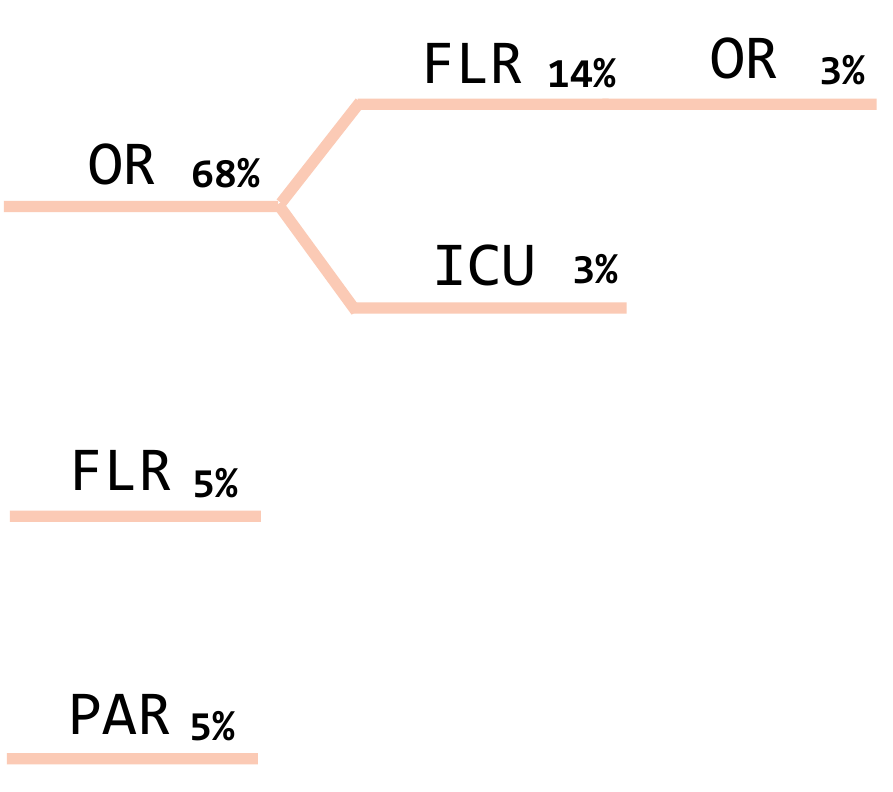}
         \caption{S2}
         \label{subfig:bedseq_2}
     \end{subfigure}
     \hfill
     \begin{subfigure}[t]{0.24\textwidth}
         \centering
         \includegraphics[width=\textwidth]{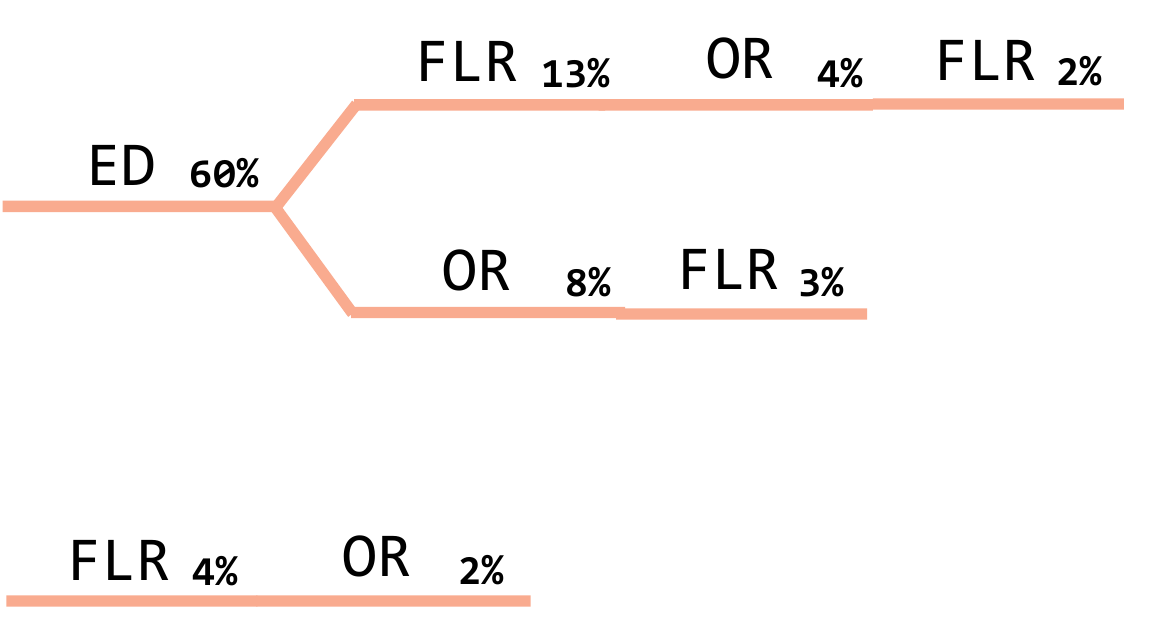}
         \caption{S3}
         \label{subfig:bedseq_3}
     \end{subfigure}
     \hfill
     \begin{subfigure}[t]{0.24\textwidth}
         \centering
         \includegraphics[width=\textwidth]{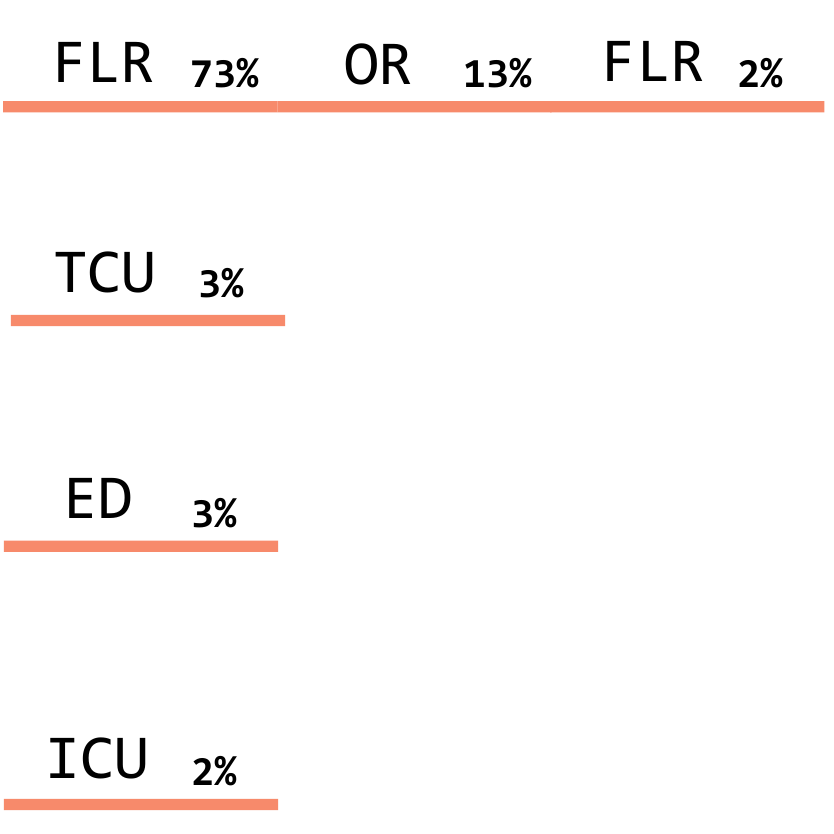}
         \caption{S4}
         \label{subfig:bedseq_4}
     \end{subfigure}
     \hfill
     \begin{subfigure}[t]{0.24\textwidth}
         \centering
         \includegraphics[width=\textwidth]{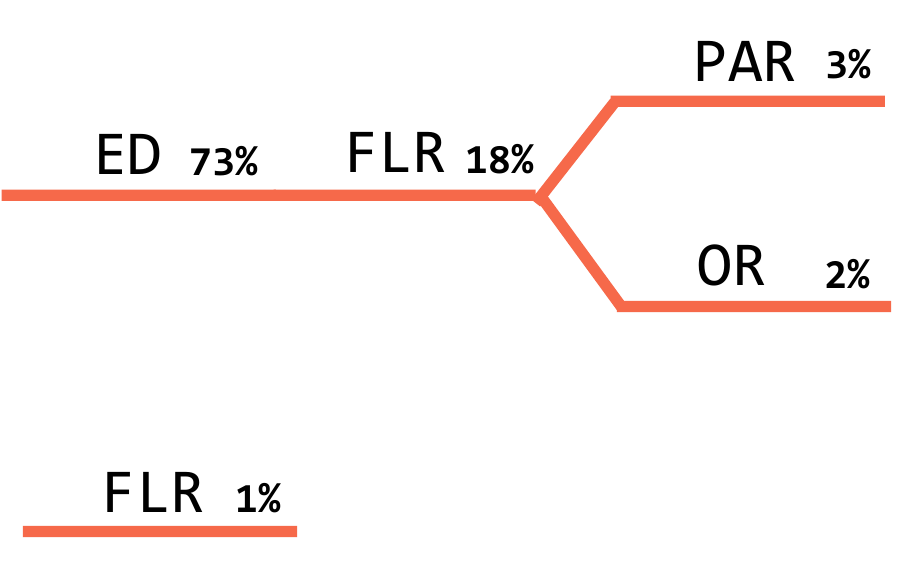}
         \caption{S5}
         \label{subfig:bedseq_5}
     \end{subfigure}
     \hfill
     \begin{subfigure}[t]{0.24\textwidth}
         \centering
         \includegraphics[width=\textwidth]{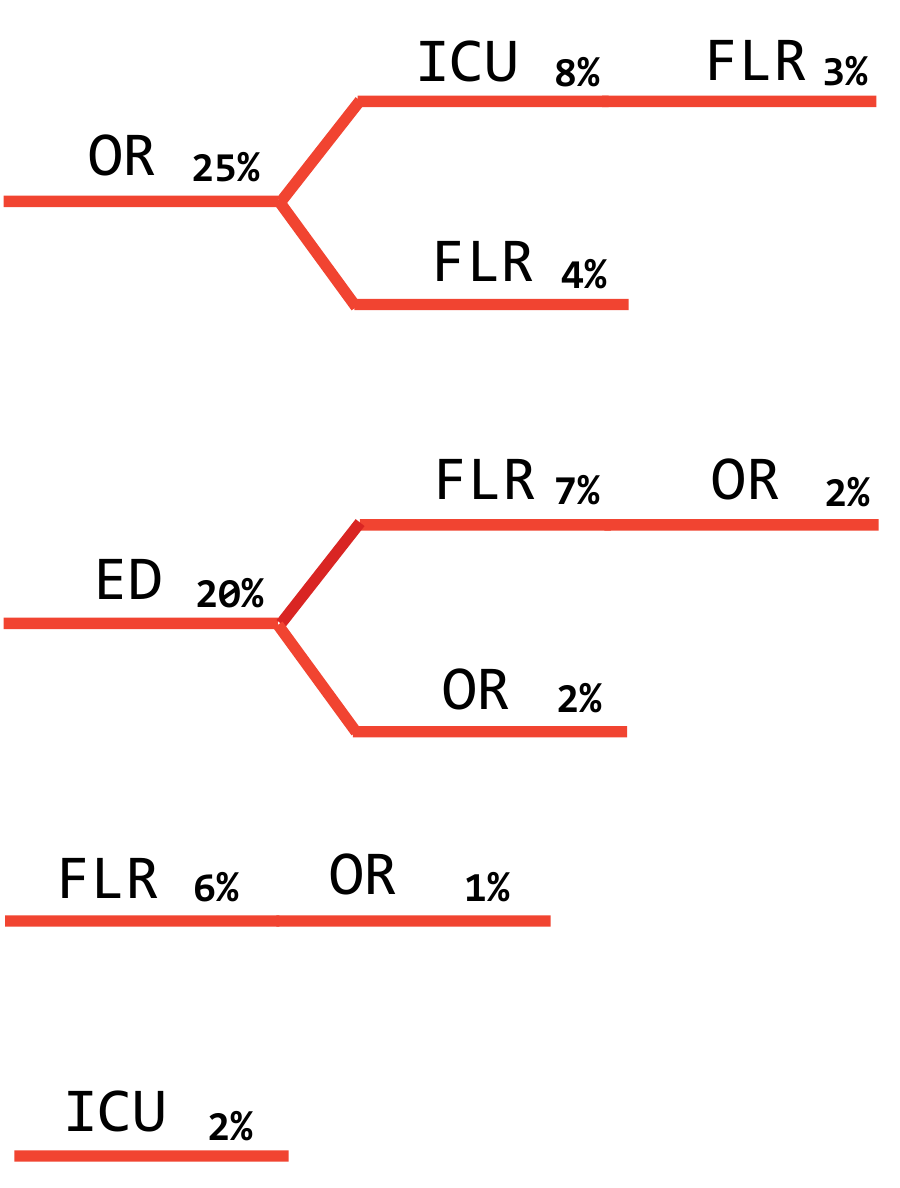}
         \caption{S6}
         \label{subfig:bedseq_6}
     \end{subfigure}
     \hfill
     \begin{subfigure}[t]{0.24\textwidth}
         \centering
         \includegraphics[width=\textwidth]{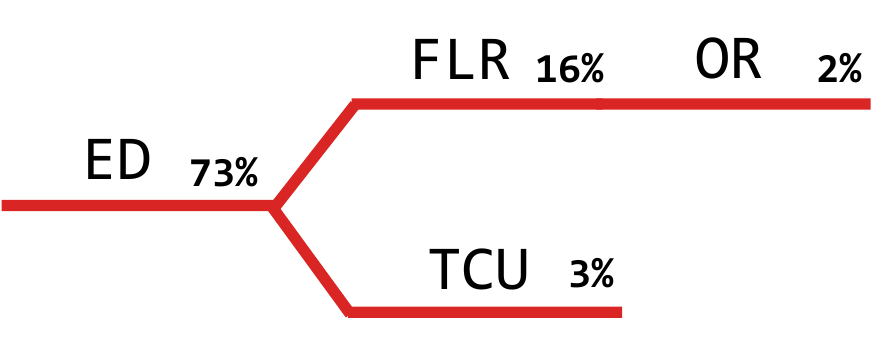}
         \caption{S7}
         \label{subfig:bedseq_7}
     \end{subfigure}
     \hfill
     \begin{subfigure}[t]{0.24\textwidth}
         \centering
         \includegraphics[width=\textwidth]{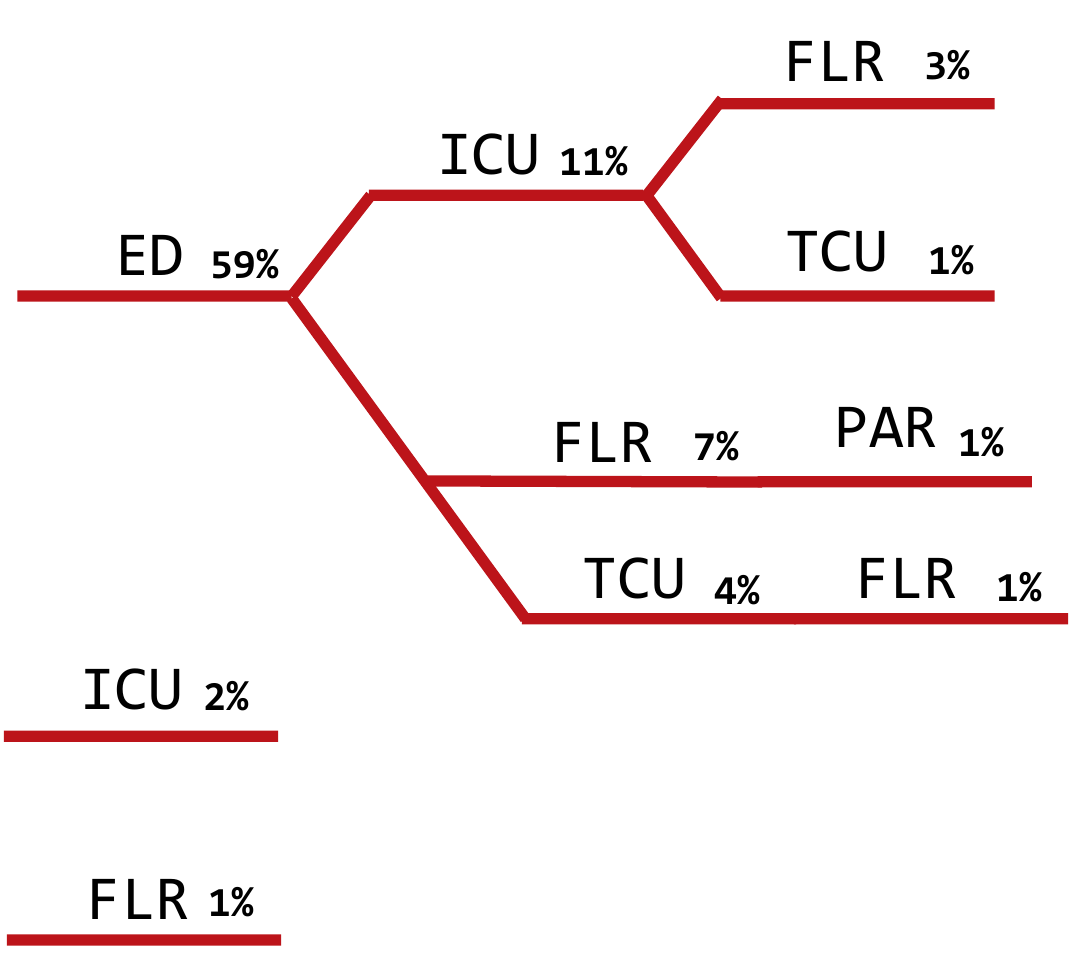}
         \caption{S8}
         \label{subfig:bedseq_8}
     \end{subfigure}
     \hfill
     \begin{subfigure}[t]{0.24\textwidth}
         \centering
         \includegraphics[width=\textwidth]{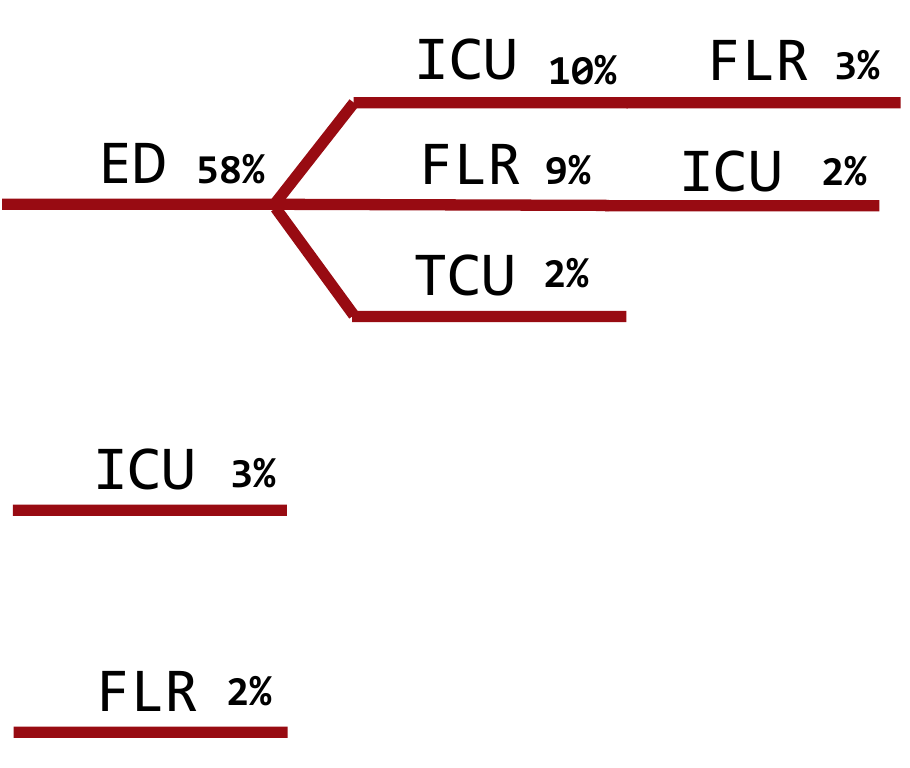}
         \caption{S9}
         \label{subfig:bedseq_9}
     \end{subfigure}
     \hfill
     \begin{subfigure}[t]{0.24\textwidth}
         \centering
         \includegraphics[width=\textwidth]{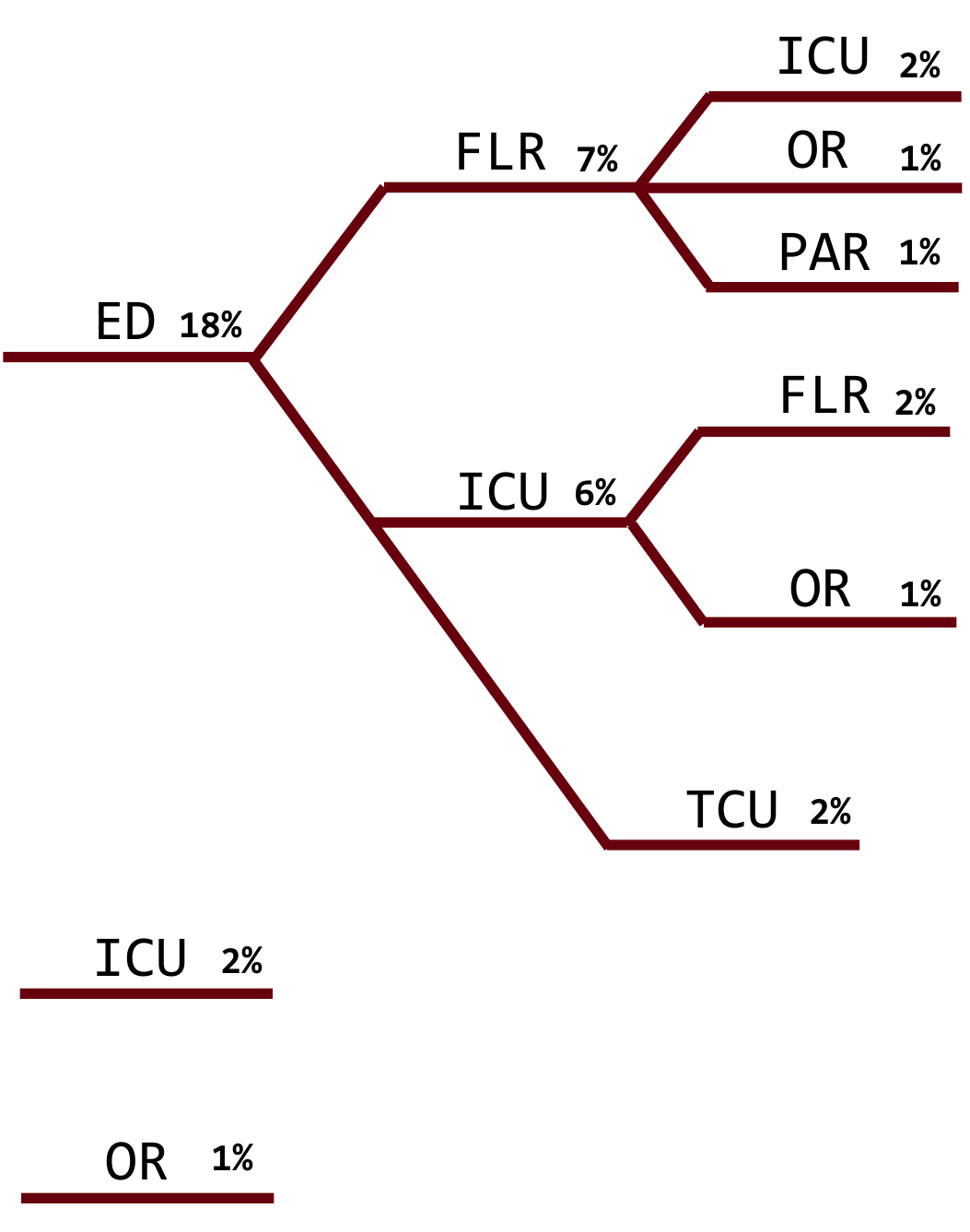}
         \caption{S10}
         \label{subfig:bedseq_10}
     \end{subfigure}
        \caption{Most likely \beds{} sequences for each of the \beds{} states. Likelihoods of the sequences are indicated by the values in the tree graphs. Only sequences with likelihood greater than or equal to 1\% are shown. The states are sorted in ascending order according to mortality enrichment.}
        \label{fig:bedseq}
\end{figure}

\beds{} sequences for each states are shown in Figure \ref{fig:bedseq}.
Each subfigure is a tree which shows the likelihood of the sequence formed by tracing from the initial bed.
This is computed by taking the initial state probabilities and forward simulating the Markov chain for each state.
The likelihood of the sequence terminating at a node in the graph is specified in the figures.
We stopped once the likelihood dropped below 1\%.
The states are sorted in ascending order according to their mortality enrichment.
%The subfigure captions include the state prevalence and the mean sequence lengths.

\subsection{\meds{} sequences}
\begin{figure}
        \centering
        \includegraphics[width=0.8\textwidth]{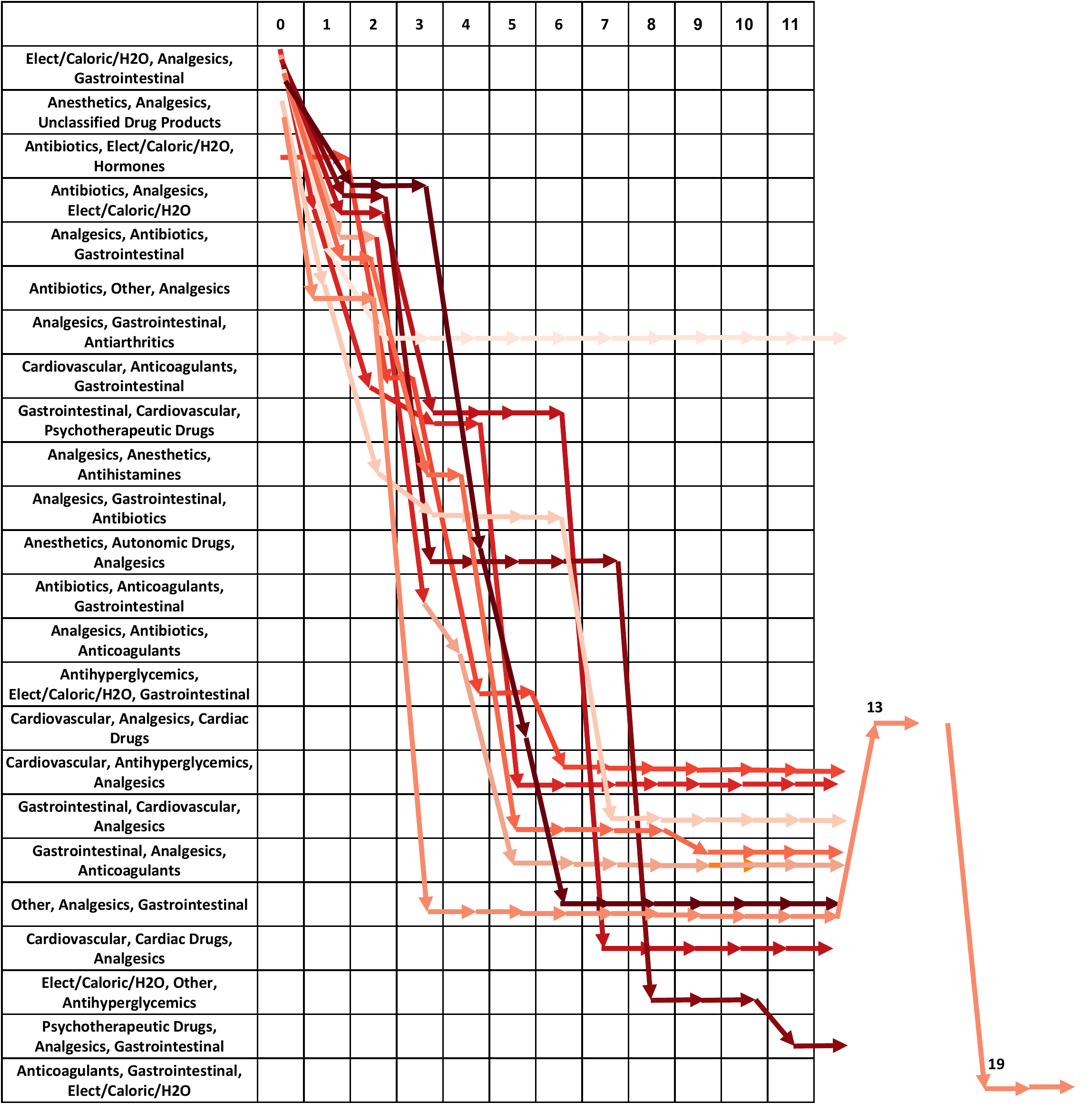}
        \caption{Most likely medication sequences for \meds{} states. Each row corresponds to an HMM state, showing the three most likely therapy classes for each state.}
        \label{fig:med_seq_all}
\end{figure}

The most likely \meds{} sequences for are shown in Figure \ref{fig:med_seq_all}.
Each row corresponds to an HMM state, with the three most likely therapy classes listed in the first column.
Each label column corresponds to a timestep, and arrows through the HMM states over time indicate the most likely sequence for each state.
%Each labeled column corresponds to a timestep, and the most likely HMM state is shown (multiple medications can be administered at each timestep).
These states are ordered roughly in order of their temporal appearance in these most likely sequences.
Colors correspond to the ordering of mortality enrichment by state.
%[include length means in the first column?]
%The complete set of \meds{} sequences are shown in Figure \ref{fig:med_seq_all}.
%\begin{figure} 
%        \centering
%        \includegraphics[width=0.8\textwidth]{figs/med_seq_few.pdf}
%        \caption{Most likely medication sequences for selected \meds{} states.}
%        \label{fig:med_seq_few}
%\end{figure}

\section{Discussion}
The model selection procedure resulted in 10 mixture states for each sequence (Table \ref{tbl:modelsel}).
For the \labs{}, \neuro{}, and \meds{} sub-models, the number of HMM states was higher (90, 70, 40, respectively).
In our results, we estimated models increasing the number of states by 10 each time.
Further granularity would be possible, at higher computational cost.

Distributions of scalar values by top-layer component are shown in Figure \ref{fig:compstat}.
The \sex{} distribution (Figure \ref{subfig:compstat_sex}) is fairly consistent around 0.5, with the exception of component 8.
\death{} varies significantly (Figure \ref{subfig:compstat_mort}) across components, with the highest probabilities in components 4, 5, and 10.
Components 6 and 8 have very low probability of \death{}.
\age{} profiles also vary (Figure \ref{subfig:compstat_age}), with component 4 having the oldest population with high probability and component 8 having the lowest \age{}.

Going a layer below these top-layer results, we examine mortality enrichment by computing the likelihood of \death{} given a sub-model state.
This results in 10 likelihoods for each sub-model (Figure \ref{tbl:enr_mort}.
States are sorted in increasing order of \death{} probability.
Note that the states are distinct across sub-models (state 1 is not the same state for \beds{} and \labs{}).
These distributions generally start near 0\% and rise to 20\% - 30\%.
Generally there is a sharp increase in risk from state S7 to S8, although this is less apparent with \neuro{}.
%A \meds{} state has the highest likelihood of \death{} (36.52\%), and the

Figure \ref{fig:enr_mort} characterizes the lowest and highest risk states for \death{} enrichment.
For \beds{}, the lowest risk state is mostly related to Obstetrics (Figure \ref{subfig:enr_mort_bed_low}).
%The highest risk state (Figure \ref{subfig:enr_mort_bed_high}) includes sequences such as \emph{ED} \rightarrow \emph{FLR} \rightarrow \emph{ICU}.
The highest risk state (Figure \ref{subfig:enr_mort_bed_high}) is likely to start at the emergency department (\emph{ED}).
The HMM sub-models (\labs{}, \neuro{}, and \meds{}) have an additional layer of latent states.
The temporal dynamics is described by likelihoods through these HMM states.
Likelihood of abnormal \neuro{} results is generally much higher for the high mortality risk state (Figure \ref{subfig:enr_mort_neuro}).
The low risk \meds{} state is most likely to proceed through two states, whereas the high risk state passes through five different states (Figure \ref{subfig:enr_mort_meds}).
Note that this is only the single most likely trajectory through the HMMs, and many other state sequences are also possible.

Figure \ref{fig:compdist} shows how the sub-model states distribute across each top-layer component.
These are ranked and color-coded by the mortality risk computed in Section \ref{sec:res_mort}.
The total probability of each component is the component probability $\alpha_z$ (same as in Figure \ref{subfig:compstat_prob}).
It can be seen that components 5 and 10 (highest \death{} probability) tend to have darker colored states.
The same is shown in Figure \ref{fig:diagdist} for \adm{} and \dis{}.

Figures \ref{fig:compstat} and \ref{fig:compdist} can be used to link sub-model states with the top layer-components.
For example, Figure \ref{subfig:compstat_sex} shows that component 8 has the highest likelihood of female \sex{}.
%Examination of component 8 reveals that this component includes Obstetrics.
Examining the \beds{} states that make up this component from Figure \ref{subfig:compdist_bed}, we see that this component includes the lowest mortality risk state (lightest color).
Figure \ref{subfig:bedseq_1} shows that \emph{OB} is a likely visit in this state.

%Figure \ref{subfig:compstat_mort} shows that components 5 and 10 have the highest likelihood of death, while components 6 and 8 have the lowest.
%To examine sequence phenotypes of \beds{}, \labs{}, and \meds{} related to these mortality risks, we first determine the states that are prevalent in these components.
%With regard to \beds{}, we can see from Figure \ref{subfig:compdist_bed} that the gray state (state 8) is attributed to component 10, but not other components.
%This state (Figure \ref{fig:bedseq}) has the longest mean sequence length of all the \beds{} states.
%\meds{} states that correspond to the high risk components are purple (state 5) and yellow (state 9).
%The trajectories for these \meds{} states can be seen in Figure \ref{fig:med_seq_all}.

%For the lowest risk, we see that the brown (state 6) and yellow (state 9) \beds{} states are prevalent (Figure \ref{subfig:compdist_bed}.

From the mean lengths shown in Figure \ref{fig:statelen}, we can see that for \labs{} and \neuro{} there is generally an increase in the mean length as the mortality risk increases.
The length of the \beds{} sequence is the number of transitions, rather than an indication of the total time in the hospital.
So, if a patient stays in one location (e.g., ICU) for a long time, it will not appear as additional items in the \beds{} sequence.
In the \meds{}, it may be the case that some mid-risk states receive a large number of \meds{} (e.g., there are clinical conditions that are not life-threatening, but still require significant extensive medication).
In all cases, the highest risk state has the largest length.
In general, the states are highly heterogeneous, and we do not necessarily expect a continuous progression of utilization as we progress from low to high mortality risk.

%Exceptions are States 7 and 9 for the \labs{}.
%The \meds{} instances are grouped by therapy classes and not individual medications, and this may be the reason why this pattern is not observed. 
%The number of total \beds{} is typically much lower and this may be why the pattern is not evident for \beds{}.
%In all cases, the highest mortality risk state has the highest length parameter.

Tables \ref{tbl:admdiagcodes} and \ref{tbl:dischdiagcodes} show the most likely (top 5) ICD codes for \adm{} and \dis{}, respectively.
The \adm{} and \dis{} sub-models share the same set of 10 diagnosis states, but only 5 distinct states for each have significant probabilities greater than 0\%.
These tables show only those 5 states that have significant probabilities.
The 5 states in Table \ref{tbl:dischdiagcodes} for \dis{} total 98.4\%, so there is a small likelihood of the other states to be in the \dis{}.
These states are sorted by mortality enrichment.
For example, the highest mortality risk \adm{} state contains Atrial fibrillation and Congestive heart failure in the top 5 codes.

Figures \ref{fig:bedseq} and \ref{fig:med_seq_all} show the likelihood of \beds{} and \meds{} sequences across states, respectively.
These are sorted and color coded by mortality risk as well.
Note that the states contain probabilistic representations and the results in Figure \ref{fig:med_seq_all} shows only the most likely trajectory through the HMM state space.
%Many of these sequences overlap across states, indicating a possible difficulty in distinguishing

%[model structure]
%While the Poisson may be more natural, we find that variance increasing with the mean is too restrictive.

%The model selection curves in Figure \ref{fig:modelsel} shows the characteristic U-shaped BIC curve.
%For a fixed $\mathcal{C}_S$, the model complexity increases with increasing $|Z|$ and therefore we expect a closer fit to the training data.
%However at the bottom of the curve, the BIC penalty increases faster than the model fit and therefore we see the BIC score increasing.

%[model exploration discussion]

\subsection{Limitations}

Our model defines the conditional independence between sequences given the top-layer latent state.
While it would be possible to extend this method to other structures that do not have this property, doing so may have detrimental consequences.
Using our model structure, we are able to control the model complexity by increasing or decreasing the state space of the top-layer latent variable.
If instead we directly model interactions between sequences, the parameter count may be too high to enable fine-tuning of the model complexity.
In addition, we may find that even second-order statistics between sequences results in a model that does not provide subgrouping properties that are desired.
However, there are many other model structures that can be explored.

As described in Section \ref{sec:data}, we model the \labs{}, but not the results of the tests.
In our modeling framework, it is possible to augment the model with additional sequences.
We decided to only model \labs{} because while it would enrich the model, our methodological development would be unaffected.
In addition, there is a level of redundancy between the \labs{}, \adm{}, \dis{}, and \meds{}.
The added benefit of including laboratory test results is an open question and application dependent.

In this work, we chose not to include vitals such as heart-rate and respiratory-rate.
Although it is possible to include vitals as another sequence, they were left out because they are often sampled at a high rate and would require significant additional model training.
In general, there are many sources of data that could be incorporated.
We had to make a decision to include enough data that can generate interesting results, while not overburdening our computational resources while developing this novel methodology.
%In addition, it would be possible to incorporate more detailed outcome assessments as well.

An aspect of the data that is not captured by our model is the explicit time-dependence.
We do incorporate sequence order for \textsf{Beds} and \textsf{Medications} sequences, but the distribution of time point is not captured.
This can be accomplished in a number of different ways, ranging from modeling the average arrival times (for example, with Exponential distributions) to full time-dependent sequence models.
Additional model complexity comes at the cost of requiring more data, and the choice should depend on the specific application at hand.

Our model is geared towards characterizing episodic sequences, rather than patient-centric modeling for following patient trajectories over time.
It assumes that each episode is drawn independently from the model, even when the same individual has 2 episodes.
Repeat episodes from the same individual may have differing characteristics from the initial episode.
Also, our data are selected from a specific timeframe, and are not guaranteed to have all episodes for any given subject.
However, the model as developed can capture differing characteristics between initial and repeat visits.
In addition, it is possible that specific mixture components are more representative of repeat visits
This analysis, and the inclusion of additional parameters to link episodes, however, is beyond the scope of this work.

\section{Conclusions}
In this work we formulated a statistical model for sequences of varying length, as commonly found in EHR data.
An important characteristic of the model is a layered set of latent variables.
The top-layer latent variable controls dependence between the sequences, and can be interpreted as a subject subgroup indicator.
Sub-model states express subgroupings in the sequence space.
Having both layers of latent variables provides an intuitive way of explored the trained model, through subgroups of subjects and subgroups of likely sequences.
The intersection of these subgroups can be examined to uncover phenotypes for specific conditions of interest.
Sequence lengths are explicitly modeled through Poisson distributions, enabling us to distinguish sequences of varying lengths.
We trained the model using EHR data and explored its internal representations through probability distributions.
%Inference algorithms were derived that compute conditional probability distributions of the sequence lengths and future presence of specific values.
%These algorithms depend crucially on the distribution of the latent variable given existing data.

The purpose of the method developed in this work is a general model capable of subgrouping and subtyping for heterogeneous sequence EHR data.
In addition to uncovering patterns in the data, this approach can be used to study disease phenotypes, since relationships between hospitalization data and certain disease states are included in the model.
To the extent that disease states are captured by diagnosis codes, this can be seen in our results relative to admission and discharge diagnosis codes (Section \ref{sec:adm_dis}, Table \ref{tbl:admdiagcodes}, and Table \ref{tbl:dischdiagcodes}).
For example, there are specific diagnosis states where Sepsis or Polyneuropathy in diabetes are most common. These could then be traced back through the top-level components to determine likely trajectories of the other sequences (e.g., \beds{}, \labs{}, etc.) that are most likely to be linked with the disease.

We perform such an analysis for mortality (Section \ref{sec:res_mort} and Figure \ref{fig:enr_mort}), where the probability of \death{} is computed for each top-layer state and related to underlying sequences sub-models.
For these analyses, however, we are using a separate variable (\death{}) to perform the analysis.
In general, the model could be augmented to incorporate variables describing a particular disease state of interest.
In that case, this analysis can be performed to learn phenotypes for the disease.

Using and applying this method is dependent on the data types being modeled and the computational resources necessary to train the model.
For sequence types that differ in format significantly from the ones we consider in this work, corresponding sub-model structures would need to be implemented.
The computational resources required are primarily for training the model.
Once trained, modest resources are required to perform subgroup analysis and inference.

As the ability to integrate heterogeneous data streams is an important aspect of precision medicine, we believe that this methodology hold promise towards the development of decision support systems.
Using data from Kaiser Permanente Northen California, we have shown how the models described can be used to subgroup patients and interpret these subgroups based on likely trajectories of data elements such as  \labs{}, \meds{}, and \beds{}.
Such patterns are often difficult to uncover because they are hidden within long sequences.
We showed how we can analyze sequences that contribute to the assessment of mortality risk.
In future work, this inference can be expanded to other areas, such as risk of disease, or length of hospital stay.
Also, since the model is a full likelihood function of the data, conditional probabilities can be computed between any subsets of the sequences to flexibly infer future sequence values.

% Acknowledgements should go at the end, before appendices and references

\section{Acknowledgments}
This work was performed under the auspices of the U.S. Department of Energy by Lawrence Livermore National Laboratory under Contract DE-AC52-07NA27344 and was supported by the LLNL LDRD Program under Project 19-ERD-009. VXL was supported in part by NIH R35GM128672.

%\newpage
\bibliographystyle{elsarticle-harv.bst}
\bibliography{refs}

\appendix
\section{Generative Description} \label{sec:app_gen}

%The generative picture hinges on an underlying and unobserved state $z$. 
%Under the model an episode $\boldsymbol{y}$ may be sampled by performing the following:
%\begin{itemize}
%    \item $z \sim Categorical\left(\boldsymbol{p}\right)$
%    \item $\phi_a \sim \mathcal{N}_q \left(\mu_z, \sigma_z^2\right)$, $\phi_s \sim Bern\left(p_{s,z}\right)$, $\phi_d \sim Bern\left(p_{d,z}\right)$
%    \item 
%\end{itemize}

The generative picture hinges on an underlying and unobserved state $z$. 
Under the model an episode $\boldsymbol{y}$ may be sampled by performing the following:
\begin{itemize}
    \item Draw a state $z$ from a categorical distribution: $z \sim Categorical\left(\boldsymbol{p}\right)$
        \item Draw an \textsf{Age} using a quantized Gaussian distribution, conditioned on the top-layer state: $\phi_a \sim \mathcal{N}_q \left(\mu_z, \sigma_z^2\right)$
        \item Draw the \textsf{Sex} using a Bernoulli distribution conditioned on the top-layer state: $\phi_s \sim Bern\left(p_{s,z}\right)$
        \item Draw the \textsf{Death} flag using a Bernoulli distribution conditioned on the top-layer state: $\phi_d \sim Bern\left(p_{d,z}\right)$
        \item Draw the \beds{} sub-model latent state from a categorical distribution, conditional on the top-layer state: $z_\beta \sim Categorical\left(\boldsymbol{p}_{\beta, z}\right)$
        \begin{itemize}
            \item Draw the number of \textsf{Beds} from a Poisson distribution, conditioned on the state: $k_\beta \sim Poisson\left(l_{\beta, z_\beta}\right)$
            \item Draw $k_\beta$ \textsf{Beds} from a Markov Chain, conditioned on the state: $\boldsymbol{\beta} \sim MarkovChain\left(\boldsymbol{p}_{\beta, z_\beta}, \boldsymbol{q}_{\beta, z_\beta}\right)$
        \end{itemize}
        \item Draw the Admission Diagnoses sub-model state from a categorical distribution, conditioned on the state: $z_\alpha \sim Categorical\left(\boldsymbol{p}_{\alpha, z}\right)$
        \begin{itemize}
            \item Draw the number of \textsf{Admission Diagnoses} from a Poisson distribution, conditioned on the state: $k_\alpha \sim Poisson\left(l_{\alpha, z_\alpha}\right)$
            \item Draw $k_\alpha$ \textsf{Admission Diagnoses} by taking repeated draws from a categorical distribution, conditioned on the state: $\boldsymbol{\alpha} \sim Categorical\left(\boldsymbol{p}_{D, z_\alpha}\right)$
        \end{itemize}
        \item Draw the Discharge Diagnoses sub-model state from a categorical distribution, conditioned on the state: $z_\delta \sim Categorical\left(\boldsymbol{p}_{\delta, z}\right)$
        \begin{itemize}
            \item Draw the number of \textsf{Discharge Diagnoses} from a Poisson distribution, conditioned on the state: $k_\delta \sim Poisson\left(l_{\delta, z_\delta}\right)$
            \item Draw $k_\delta$ \textsf{Discharge Diagnoses} by taking repeated draws from a categorical distribution, conditioned on the state: $\boldsymbol{\delta} \sim Categorical\left(\boldsymbol{p}_{D, z_\delta}\right)$
        \end{itemize}
        
        \item Draw the \labs{} sub-model state from a categorical distribution, conditioned on the state: $z_\lambda \sim Categorical\left(\boldsymbol{p}_{\lambda, z}\right)$
        \begin{itemize}
            \item Draw the number of distinct \labs{} instances from a Poisson distribution, conditioned on the state: $k_\lambda \sim Poisson\left(l_{\lambda, z_\lambda}\right)$
            \item Draw a \labs{} HMM state sequence from a Markov Chain: $\boldsymbol{s}_\lambda \sim MarkovChain\left(\boldsymbol{p}_{\lambda, z_\lambda}, \boldsymbol{q}_{\lambda, z_\lambda}\right)$
            \item Draw the number of \labs{} given for each instance from a Poisson distribution, conditioned on the state: $k_{\lambda, s_\lambda} \sim Poisson\left(l_{L, s_\lambda}\right)$
            \item Draw a set of $k_{\lambda, s_\lambda}$ \textsf{Laboratory Tests} for each instance by taking repeated draws of a Categorical distribution, conditioned on the state: $\boldsymbol{\lambda}_{s_\lambda} \sim Categorical\left(\boldsymbol{p}_{L, s_\lambda}\right)$
        \end{itemize}
        
        \item Draw the \neuro{} sub-model state from a categorical distribution, conditioned on the state: $z_\nu \sim Categorical\left(\boldsymbol{p}_{\nu, z}\right)$
        \begin{itemize}
            \item Draw the number of distinct \neuro{} instances from a Poisson distribution, conditioned on the state: $k_\nu \sim Poisson\left(l_{\nu, z_\nu}\right)$
            \item Draw a \neuro{} HMM state sequence from a Markov Chain: $\boldsymbol{s}_\nu \sim MarkovChain\left(\boldsymbol{p}_{\nu, z_\nu}, \boldsymbol{q}_{\nu, z}\right)$
            \item Draw the number of \neuro{} given for each instance from a Poisson distribution, conditioned on the state: $k_{\nu, s_\nu} \sim Poisson\left(l_{N, s_\nu}\right)$
            \item Draw a set of $k_{\nu, s_\nu}$ \neuro{} for each instance by taking repeated draws of a Categorical distribution, conditioned on the state: $\boldsymbol{\nu}_{s_\nu} \sim Categorical\left(\boldsymbol{p}_{N, s_\nu}\right)$
        \end{itemize}
        
        \item Draw the \meds{} sub-model state from a categorical distribution, conditioned on the state: $z_\mu \sim Categorical\left(\boldsymbol{p}_{\mu, z}\right)$
        \begin{itemize}
            \item Draw the number of distinct \meds{} instances from a Poisson distribution, conditioned on the state: $k_\mu \sim Poisson\left(l_{\mu, z_\mu}\right)$
            \item Draw a \meds{} HMM state sequence from a Markov Chain: $\boldsymbol{s}_\mu \sim MarkovChain\left(\boldsymbol{p}_{\mu, z_\mu}, \boldsymbol{q}_{\mu, z_\mu}\right)$
            \item Draw the number of \textsf{Medications} given for each instance from a Poisson distribution, conditioned on the state: $k_{\mu, s_\mu} \sim Poisson\left(l_{M, s_\mu}\right)$
            \item Draw a set of $k_{\mu, s_\mu}$ \textsf{Medications} for each instance by taking repeated draws of a Categorical distribution, conditioned on the state: $\boldsymbol{\mu}_s \sim Categorical\left(\boldsymbol{p}_{M, s_\mu}\right)$
        \end{itemize}
\end{itemize}

\section{Probability Density Functions} \label{sec:app_pdf}

For \age{}, we use a quantized and truncated Gaussian distribution.
This conditional distribution is,
    $$
        f\left(\phi_a | Z=z\right) = \frac{1}{C\sqrt{2\pi\sigma_z^2}}e^{\frac{\left(m_z - \phi_a\right)^2}{2\sigma_z^2}},
    $$
where the normalizing value ensures that the distribution sums to 1, $C=\sum_{\phi_a}\frac{1}{\sqrt{2\pi\sigma_z^2}}e^{\frac{\left(m_z - \phi_a\right)^2}{2\sigma_z^2}}$.

For \sex{} and \death{}, we we use Bernoulli distributions,
    $$
        f\left(\phi_s | Z=z\right) = p_{s,z}^{1 - \phi_s}
        \left(1 - p_{s,z}\right)^{\phi_s},
    $$
    $$
        f\left(\phi_d | Z=z\right) = p_{d,z}^{1 - \phi_d}
        \left(1 - p_{d,z}\right)^{\phi_d}.
    $$
    
For the \beds{} sequence we use a mixture of Markov chains, where the length of the sequence is captured by  Poisson distribution.
The distribution is,
    $$
        f\left(\boldsymbol{\beta} | Z=z\right) = 
        \sum_{z_\beta}
        p_{z, z_\beta}
        \frac{l_{z_\beta}^{|\boldsymbol{\beta}|}}{|\boldsymbol{\beta}|!}e^{-l_{\beta, z_\beta}}
        p_{\beta,z_\beta, \beta_1} \prod_{i=1}^{|\boldsymbol{\beta}| - 1} q_{\beta, z_\beta, \beta_i, \beta_{i+1}},
    $$
where $p_{\beta,z_\beta,i}$ is the probability that the first item is $i$ conditioned on $Z_\beta=z_\beta$, and $q_{\beta, z_\beta, i, j}$ is the probability of transitioning from item $i$ to $j$ conditioned on $Z\beta=z_\beta$.
    
For \adm{} and  \dis{} we use mixtures over products of categorical distributions.
The length of each of these sequences for the entire episode ($|\boldsymbol{\alpha}|$ for \adm{} and $|\boldsymbol{\delta}|$ for \dis{}) is modeled using a Poisson distribution.
For these sets we have,

    $$
        f\left(\boldsymbol{\alpha} | Z=z\right) = 
        \sum_{z_\alpha}
        p_{z, z_\alpha}
        \frac{l_{\alpha, z_\alpha}^{|\boldsymbol{\alpha}|}}{|\boldsymbol{\alpha}|!}e^{-l_{\alpha, z_\alpha}}
        \prod_{i=1}^{|\boldsymbol{\alpha}|} p_{\alpha, z_\alpha, \alpha_i},
    $$
    $$
        f\left(\boldsymbol{\delta} | Z=z\right) = 
        \sum_{z_\delta}
        p_{z, z_\delta}
        \frac{l_{\delta, z_\delta}^{|\boldsymbol{\delta}|}}{|\boldsymbol{\delta}|!}e^{-l_{\delta, z_\delta}}
        \prod_{i=1}^{|\boldsymbol{\delta}|} p_{\delta, z_\delta, \delta_i},
    $$
    %$$
    %    f\left(\boldsymbol{\lambda} | Z=z\right) = 
    %    \frac{l_{\lambda, z}^{\sum_{i=1}^{|\boldsymbol{\lambda}|} |\boldsymbol{\lambda}_i|}}{\sum_{i=1}^{|\boldsymbol{\lambda}|} |\boldsymbol{\lambda}_i|!}e^{-l_{\lambda, z}}
    %    \prod_{i=1}^{|\boldsymbol{\lambda}|} 
    %    \prod_{j=1}^{|\boldsymbol{\lambda}_i|}
    %    p_{\lambda, z, \lambda_{i,j}},
    %$$
%
    %$$
    %    f\left(\boldsymbol{\nu} | Z=z\right) = 
    %    \frac{l_{\nu, z}^{\sum_{i=1}^{|\boldsymbol{\nu}|} |\boldsymbol{\nu}_i|}}{\sum_{i=1}^{|\boldsymbol{\nu}|} |\boldsymbol{\nu}_i|!}e^{-l_{\nu, z}}
    %    \prod_{i=1}^{|\boldsymbol{\nu}|} 
    %    \prod_{j=1}^{|\boldsymbol{\nu}_i|}
    %    p_{\nu, z, \nu_{i,j}},
    %$$
%where $N_{\cdot, i}$ is the number of occurrences of item $i$ in the $\cdot$ sequence and $|\mathcal{C_\cdot}|$ is the number of possible items in the $\cdot$ sequence.
where $p_{x, z_\alpha, i}$ is the probability of item $i$ in sequence $x$ for sub-model latent state $z_\alpha$, and $p_{x, z_\delta, i}$ is the probability of item $i$ in sequence $x$ for sub-model latent state $z_\delta$

The \labs{}, \neuro{}, and \meds{} sub-models can have multiple items at any timepoint.
%We also want the model to capture the order and transitions between timepoints in the \textsf{Medications} sequence.
%However, unlike the \textsf{Beds} sequence, there can be multiple items at any timepoint.
%A Markov Chain is not sufficient to characterize this feature of the data.
This is modeled using HMMs, where where the HMM transitions there a latent state sequence, $\boldsymbol{S}$, over time.
Each state can have any number of observations, drawn from a categorical distribution.
%In order to allow for this feature, we utilize an extension to the Markov Chain model.
%We use a sequence of latent states, $\boldsymbol{S}$, where each of these states carries a conditional distribution over a collection of medications of arbitrary length.
The number of observations is captured with a Poisson distribution conditioned on the state.
In this way we are able to model the ordered sequence and express multiple simultaneous observations.
A mixture of these HMM models is used for the \labs{}, \neuro{}, and \meds{} observations.
%The result is a Hidden Markov Model with variable length observations for each state, $s$, characterized by Poisson distributions.
This distributions for these are,
    $$
        f\left(\boldsymbol{\lambda} | Z=z\right) = 
        \sum_{z_\lambda}
        p_{z, z_\lambda}
        \frac{l_{\lambda, z_\lambda}^{|\boldsymbol{\lambda}|}}{|\boldsymbol{\lambda}|!}e^{-l_{\lambda, z_\lambda}}
        \sum_{\boldsymbol{S}}
        \left(
        p_{\lambda,z_\lambda,s_1} \prod_{i=1}^{|\boldsymbol{\lambda}| - 1} q_{\lambda,z_\lambda, s_i, s_{i+1}}
        \right)
        \left(
        \prod_{i=1}^{|\boldsymbol{\lambda}|}
        \frac{l_{\lambda, z_\lambda, s_i}^{|\boldsymbol{\lambda}_i|}}{|\boldsymbol{\lambda}_i|!}e^{-l_{\lambda, z_\lambda, s_i}}
        \prod_{j=1}^{|\boldsymbol{\lambda}_i|} p_{\lambda, z_\lambda, s_i, \lambda{i, j}}
        \right),
    $$
        $$
        f\left(\boldsymbol{\nu} | Z=z\right) = 
        \sum_{z_\nu}
        p_{z, z_\nu}
        \frac{l_{\nu, z_\nu}^{|\boldsymbol{\nu}|}}{|\boldsymbol{\nu}|!}e^{-l_{\nu, z_\nu}}
        \sum_{\boldsymbol{S}}
        \left(
        p_{\nu,z_\nu,s_1} \prod_{i=1}^{|\boldsymbol{\nu}| - 1} q_{\nu,z_\nu, s_i, s_{i+1}}
        \right)
        \left(
        \prod_{i=1}^{|\boldsymbol{\nu}|}
        \frac{l_{\nu, z_\nu, s_i}^{|\boldsymbol{\nu}_i|}}{|\boldsymbol{\nu}_i|!}e^{-l_{\nu, z_\nu, s_i}}
        \prod_{j=1}^{|\boldsymbol{\nu}_i|} p_{\nu, z_\nu, s_i, \nu{i, j}}
        \right),
    $$
        $$
        f\left(\boldsymbol{\mu} | Z=z\right) = 
        \sum_{z_\mu}
        p_{z, z_\mu}
        \frac{l_{\mu, z_\mu}^{|\boldsymbol{\mu}|}}{|\boldsymbol{\mu}|!}e^{-l_{\mu, z_\mu}}
        \sum_{\boldsymbol{S}}
        \left(
        p_{\mu,z_\mu,s_1} \prod_{i=1}^{|\boldsymbol{\mu}| - 1} q_{\mu,z_\mu, s_i, s_{i+1}}
        \right)
        \left(
        \prod_{i=1}^{|\boldsymbol{\mu}|}
        \frac{l_{\mu, z_\mu, s_i}^{|\boldsymbol{\mu}_i|}}{|\boldsymbol{\mu}_i|!}e^{-l_{\mu, z_\mu, s_i}}
        \prod_{j=1}^{|\boldsymbol{\mu}_i|} p_{\mu, z_\mu, s_i, \mu_{i, j}}
        \right),
    $$
    
where $p_{z, z_\lambda}$ is the probability of sub-model state $z_\lambda$ given the top-layer state $z$, $p_{\lambda, z_\lambda, i}$ is the probability that state $i$ is the initial state, $q_{\lambda, z_\lambda, i, j}$ is the transition probability from HMM state $i$ to HMM state $j$, and $p_{\lambda, z_\lambda, i, j}$ is the probability of medication $j$ conditioned on HMM state $i$ and sub-model state $z_\lambda$.
And the same holds (substituting $\nu$ or $\mu$ for $\lambda$) for the \neuro{} and \meds{} models.

The complete set of parameters is shown in Table \ref{tbl:params}, along with the number of parameters in each category.
In this Table, $|Z|$ is the number of top-layer latent states, $|Z_x|$ is the number of sub-model latent states for model $x$, $\mathcal{C}_x$ is the number of possible items in $x$.
The number of HMM states is $\mathcal{C}_{S,x}$ for HMM model $x$.
    
%\begin{table}[ht]
%    \centering
%     \begin{tabular}{|c c c|} 
%     \hline
%     Category & Parameters & Number \\ [0.5ex] 
%     \hline
%     Latent variable & $p_z$ & $|Z|$ \\
%     Quantized Gaussian  & $m_z, \sigma_z^2$ & $2|Z|$ \\
%     Bernoulli  & $p_{s,z}, p_{d,z}$ & 2 \\
%     Sequence length  & $l_{a,z}, l_{d,z}, l_{\lambda, z}, l_{\nu, z}, l_{\beta, z}, l_{\mu, z}$ & $6|Z|$ \\
%     Categorical &  $p_{\alpha, z, i}, p_{\delta, z, i}, p_{\lambda, z, i}, p_{\nu, z, i}$ & $\left(\mathcal{C}_\alpha + \mathcal{C}_\delta + \mathcal{C}_\lambda + \mathcal{C}_\nu\right)|Z|$ \\
%     Markov chain & $p_{\beta, z, i}, q_{\beta, z, i, j}, p_{\mu, z, i}, q_{\mu, z, i, j}$ & $\left(\mathcal{C}_\beta + \mathcal{C}_\beta^2 + \mathcal{C}_S + \mathcal{C}_S^2\right)|Z|$ \\
%     Hidden Markov Model & $p_{\mu, z, s, i}$ & $\mathcal{C}_S\mathcal{C}_\mu|Z|$ \\
%     \hline
%     \end{tabular}
%     \caption{Model Parameters}
%     \label{tbl:params}
%\end{table}
\begin{table}[ht]
    \centering
     \begin{tabular}{|c c c|} 
     \hline
     Data Element & Parameters & Count \\ [0.5ex] 
     \hline
     Top-layer latent state $Z$         & $p_z$                                                             & $|Z|$ \\
     \adm{} latent state $Z_\alpha$     & $p_{z, z_\alpha}$                                                 & $|Z| \times |Z_\alpha|$ \\
     \dis{} latent state $Z_\delta$     & $p_{z, z_\delta}$                                                 & $|Z| \times |Z_\delta|$ \\
     \beds{} latent state $Z_\beta$     & $p_{z, z_\beta}$                                                  & $|Z| \times |Z_\beta|$ \\
     \labs{} latent state $Z_\lambda$   & $p_{z, z_\lambda}$                                                & $|Z| \times |Z_\lambda|$ \\
     \neuro{} latent state $Z_\nu$      & $p_{z, z_\nu}$                                                    & $|Z| \times |Z_\nu|$ \\
     \meds{} latent state $Z_\mu$       & $p_{z, z_\mu}$                                                    & $|Z| \times |Z_\mu|$ \\
     \age{}                             & $m_z, \sigma_z^2$                                                 & $2|Z|$ \\
     \sex{}                             & $p_{s, z}$                                                        & $|Z|$ \\
     \death{}                           & $p_{d,z}$                                                         & $|Z|$ \\
     \adm{}                             & $l_{\alpha,z_\alpha}, p_{\alpha, z_\alpha, i}$                                  & $\left(1 + \mathcal{C}_\alpha\right)|Z_\alpha|$ \\
     \dis{}                             & $l_{\delta,z_\delta}, p_{\delta, z_\delta, i}$                                  & $\left(1 + \mathcal{C}_\delta\right)|Z_\delta|$ \\
     \beds{}                            & $l_{\beta, z_\beta}, p_{\beta, z_\beta, i}, q_{\beta, z_\beta, i, j}$               & $\left(1 + \mathcal{C}_\beta + \mathcal{C}_\beta^2\right)|Z_\beta|$ \\
     %\textsf{Laboratory Tests}          & $l_{\lambda, z}, p_{\lambda, z, i}$                               & $\left(1 + \mathcal{C}_\lambda\right)|Z|$ \\
     %\textsf{Neurological Tests}        & $l_{\nu, z}, p_{\nu, z, i}$                                       & $\left(1 + \mathcal{C}_\nu\right)|Z|$ \\
     \labs{}                            & $l_{\lambda, z_\lambda}, p_{\lambda, z_\lambda, i}, q_{\lambda, z_\lambda, i, j}, p_{\lambda, z_\lambda, s, i}$   & $\left(1 + \mathcal{C}_{S, \lambda} + \mathcal{C}_{S,\lambda}^2 + \mathcal{C}_{S, \lambda}\mathcal{C}_\lambda\right)|Z_\lambda|$ \\
     \neuro{}                           & $l_{\nu, z_\nu}, p_{\nu, z_\nu, i}, q_{\nu, z_\nu, i, j}, p_{\nu, z_\nu, s, i}$   & $\left(1 + \mathcal{C}_{S, \nu} + \mathcal{C}_{S,\nu}^2 + \mathcal{C}_{S, \nu}\mathcal{C}_\nu\right)|Z_\nu|$ \\
     \meds{}                             & $l_{\mu, z_\mu}, p_{\mu, z_\mu, i}, q_{\mu, z_\mu, i, j}, p_{\mu, z_\mu, s, i}$   & $\left(1 + \mathcal{C}_{S, \mu} + \mathcal{C}_{S,\mu}^2 + \mathcal{C}_{S, \mu}\mathcal{C}_\mu\right)|Z_\mu|$ \\
     \hline
     \end{tabular}
     \caption{Model parameters for each data element and the parameter count.}
     \label{tbl:params}
\end{table}

\end{document}